\documentclass[journal]{IEEEtran}

\usepackage{mathrsfs}
\usepackage{amssymb}
\usepackage[cmex10]{amsmath}
\usepackage{amsthm}
\usepackage{graphics}
\usepackage{latexsym}
\usepackage{amsfonts}
\usepackage{color}
\usepackage{ctable}
\usepackage{graphicx}
\usepackage{verbatim}
\usepackage{epstopdf}
\usepackage{dsfont}

\ifCLASSOPTIONcompsoc
\usepackage[tight,normalsize,sf,SF,center]{subfigure}
\else
\usepackage[tight,footnotesize,center]{subfigure}
\fi
\newtheorem{theo}{Theorem}
\newtheorem{lemm}[theo]{Lemma}
\newtheorem{coro}[theo]{Corollary}
\newtheorem{prop}[theo]{Proposition}
\newtheorem{defi}[theo]{Definition}

\theoremstyle{remark}
\newtheorem{exam}{Example}
\newtheorem{rema}[theo]{Remark}

\newcommand{\be}{\begin{IEEEeqnarray*}{rCl}}
\newcommand{\ee}{\end{IEEEeqnarray*}}
\newcommand{\ben}{\begin{IEEEeqnarray}{rCl}}
\newcommand{\een}{\end{IEEEeqnarray}}
\newcommand{\lb}[1]{\left[\begin{array}{#1}}
\newcommand{\rb}{\end{array}\right]}
\newcommand{\lp}[1]{\left(\begin{array}{#1}}
\newcommand{\rp}{\end{array}\right)}

\newcommand{\leftd}[1]{\left\{\begin{array}{#1}}
\newcommand{\rightd}{\end{array}\right.}

\def\A {\mathbf{A}}
\def\B {\mathbf{B}}
\def\C {\mathbf{C}}
\def\D {\mathbf{D}}

\def\G {\mathbf{G}}
\def\H {\mathbf{H}}
\def\I {\mathbf{I}}

\def\K {\mathbf{K}}
\def\L {\mathbf{L}}
\def\M {\mathbf{M}}

\def\P {\mathbf{P}}
\def\Q {\mathbf{Q}}
\def\R {\mathbf{R}}

\def\T {\mathbf{T}}
\def\U {\mathbf{U}}

\def\W {\mathbf{W}}
\def\X {\mathbf{X}}
\def\Y {\mathbf{Y}}

\def\a {\mathbf{a}}
\def\b {\mathbf{b}}
\def\cc {\mathbf{c}}

\def\e {\mathbf{e}}

\def\h{\mathbf{h}}
\def\hh{\mathbf{h}}

\def\m {\mathbf{m}}

\def\s {\mathbf{s}}
\def\t {\mathbf{t}}

\def\v {\mathbf{v}}
\def\w {\mathbf{w}}
\def\x {\mathbf{x}}
\def\y {\mathbf{y}}
\def\z {\mathbf{z}}

\def\Bs {\mathscr{B}}

\def\Ds {\mathscr{D}}

\def\Bc {\mathcal{B}}

\def\Fc {\mathcal{F}}

\def\Hc {\mathcal{H}}

\def\Jc {\mathcal{J}}
\def\Kc {\mathcal{K}}

\def\Mc {\mathcal{M}}
\def\Nc {\mathcal{N}}

\def\Sc {\mathcal{S}}

\def\Lbd{\boldsymbol{\Lambda}}
\def\PSI {\boldsymbol{\psi}}

\def\muu{\boldsymbol{\mu}}
\def\teta{\boldsymbol{\theta}}
\def\Dta{\boldsymbol{\Delta}}

\def\Rb {\mathbb{R}}
\def\Pb {\mathbb{P}}
\def\Eb {\mathbb{E}}

\def\Ob {\mathbb{O}}

\def\gg{\boldsymbol{g}}

\newcommand{\sign}{\mathrm{sign}}

\newcommand{\cas}{\xrightarrow[N\rightarrow\infty]{a.s.}}
\newcommand{\CD}{\xrightarrow[N\rightarrow\infty]{\Ds}}
\newcommand{\diag}{\mathrm{diag}}
\newcommand{\defeq}{\stackrel{\mathrm{def}}{=}}
\newcommand{\TR}{\mathsf{T}}
\newcommand{\argmin}[1]{\underset{#1}{\operatorname{argmin\,\,}}}

\newcommand{\cov}{\mathrm{Cov}}

\newcommand{\EE}{{\Eb}_N}

\newcommand{\hatfc}{\widehat{\Fc}}
\newcommand{\hatw}{\widehat{\W}}
\newcommand{\hatc}{\widehat{\C}}
\newcommand{\hata}{\widehat{\A}}
\newcommand{\hatb}{\widehat{\B}}

\newenvironment{Tabular}[2][1]
  {\def\arraystretch{#1}\tabular{#2}}
  {\endtabular}
\newcounter{mytempeqncnt}

\title{A convergence and asymptotic analysis of the generalized symmetric FastICA algorithm}
\author{Tianwen~Wei,~\IEEEmembership{Member,~IEEE}
\thanks{Copyright (c) 2015 IEEE. Personal use of this material is permitted. However, permission to use this material for any other purposes must be obtained from the IEEE by sending a request to pubs-permissions@ieee.org.
}
\thanks{
T. Wei was with
Laboratoire de Math\'ematiques de Besan\c{c}on, University of Franche-Comt\'e, Besan\c{c}on, France. He is now with the Department of Statistics and Mathematics, Zhongnan University of Economics and Law, Wuhan, China.
Corresponding Author: tianwen.wei.2014@ieee.org.
Part of this work was presented \cite{WEISSP2} at IEEE workshop on Statistical Signal Processing 2014, Gold Coast, Australia.
}
}
\begin{document}
\maketitle

\begin{abstract}
This contribution deals with the FastICA algorithm in the domain of Independent Component Analysis (ICA).
The focus is on the asymptotic behavior of the generalized symmetric variant of the algorithm. The latter has already been shown to possess the potential to achieve the Cram\'er-Rao Bound (CRB) by allowing the usage of  different nonlinearity functions in its implementation.
Although the FastICA algorithm along with its variants are among the most extensively studied methods in the domain of ICA,
 a rigorous study of the asymptotic distribution of the
 generalized symmetric FastICA algorithm is still missing.
 In fact, all the existing results exhibit certain limitations. Some ignores the impact of data standardization on the asymptotic
 statistics; others are only based on heuristic arguments.
 In this work, we aim at deriving general and rigorous results on the limiting distribution and the asymptotic statistics of the FastICA algorithm.
 We begin by showing that the generalized symmetric FastICA optimizes a function that is a sum of the contrast functions of traditional one-unit FastICA with a correction of the sign. Based on this characterization, we established the asymptotic normality and derived a closed-form analytic expression of the asymptotic covariance matrix of the generalized symmetric FastICA estimator using the method of estimating equation and M-estimator. Computer simulations are also provided, which support the theoretical results.
\end{abstract}

\section{Introduction}
\IEEEPARstart{T}{he}  {Independent Component Analysis (ICA)} is a general framework for solving Blind Source Separation (BSS) problems \cite{HYVABOOK,COMOBOOK,AMABOOK}. It
is a statistical and computational method which aims at extracting the unobserved source signals from their linear mixtures.
 The fundamental assumption of ICA is that the source signals are statistically independent.
Up to date, there exist various ICA algorithms \cite{CARD1993, COMO94, RADICAL, ZARZ2010} in the community, see \cite{COMOBOOK} for more details. One of the most widely used ICA algorithms is the FastICA algorithm, proposed by Hyv\"arinen and Oja \cite{HYVABOOK, HYVAOJA97, HYVA99}. It is based on the optimization of
a contrast function that measures the non-Gaussianity of the mixture.
The popularity of FastICA can be attributed to its simplicity, ease of implementation, and flexibility to choose the nonlinearity function.

Among many variants, there are two basic versions of FastICA: the one-unit (or deflation) FastICA and the symmetric FastICA.
The one-unit version of FastICA corresponds to the sequential (or deflationary) source separation scheme: it extracts one source at a time until all the sources are recovered. To avoid that the algorithm converges to the same source twice, an additional deflationary procedure is required \cite{DELF}.
This version of FastICA suffers the common drawback of all sequential source separation scheme: the error propagation during successive extraction for problems with large dimensionality \cite{OLLI, MIET2014a}.  The symmetric FastICA \cite{OJAYUAN} corresponds to the simultaneous (or parallel, symmetric) source separation scheme:
all the source signals are to be extracted simultaneously. It can be considered as several one-unit FastICA implemented in parallel, with the projection step replaced by a matrix orthonormalization in each iteration.
Symmetric FastICA is shown to be more stable and reliable in practice, at the cost of relatively high computation load compared to the one-unit version of FastICA.
The generalized symmetric FastICA algorithm \cite{TICHOJA2} is a generalization of the ordinary symmetric version of FastICA. It features the usage of possibly different nonlinearity functions in its parallel implementations of one-unit FastICA. This is motivated by the discovery of the Cram\'er-Rao Bound (CRB) \cite{TICHOJA,OLLI2008} of the linear ICA. It is shown \cite{TICHOJA2} that if the nonlinearity functions are adapted to the distributions of the sources, then the CRB can be achieved as the sample size tends to infinity. In other words, the generalized FastICA algorithm with the optimal nonlinearity functions can be asymptotically efficient.

The FastICA algorithm has been extensively studied since its invention.
It is shown to possess locally at least quadratic convergence speed \cite{HYVA99, OJAYUAN, SHEN}, and the convergence is monotonic \cite{REGA}. In particular, for kurtosis-based FastICA,
it is proved that there does not exist spurious fixed points \cite{DOUG2003} and the convergence speed becomes cubic \cite{HYVA99}.
 FastICA has also been generalized to cope with complex valued signals \cite{HYVABING}.


This paper studies the generalized symmetric FastICA algorithm, with the focus on deriving the asymptotic covariance matrix.
Although the asymptotic performance of the FastICA algorithm has already been studied by many authors, no special attention was given to the generalized symmetric version of the algorithm. Besides, all the existing results exhibit certain limitations: some are based on a heuristic approach \cite{HYVA97, TICHOJA, OLLI}; many only deal with the one-unit version of FastICA \cite{HYVA97,OLLI, OLLI2011, REYH2012, WEI2, MIET2014a}; most importantly, the majority overlooks the impact of data centering or data whitening on the asymptotic behavior of the algorithm \cite{HYVA97, TICHOJA, HYVA2006}.
We give a detailed review of the literature in Section \ref{review}.
Another interest of studying the asymptotic performance of the generalized FastICA stems from the claim \cite{TICHOJA2} that it has the potential to achieve the CRB, which is based on the
expression of the asymptotic covariance matrix of symmetric FastICA
  derived in \cite{TICHOJA}. However, this claim is questionable. In fact, as is noticed in \cite{OLLI}, the expression given in \cite{TICHOJA} is flawed: it is valid only if the underlying sources have symmetric distributions. Therefore, it is vital to derive the correct expression for the general case and check whether or not the CRB is still attainable. In this contribution, we shall eventually give a positive response to this question.

 We organize this work as follows. In Section II, we define the  basic notions of linear ICA, e.g. data model, data standardization, demixing matrix, etc.
 Section III aims at introducing three variants of the FastICA algorithm, namely one-unit FastICA, symmetric FastICA and generalized symmetric FastICA. In Section IV, we characterize the fixed points of the symmetric FastICA algorithm, showing that they are local optimizers of a sum of the contrast functions used by traditional one-unit FastICA with a correction of the sign.
This result reveals a link to the method of estimating equation and M-estimator. In Section V, we establish the asymptotic normality of the generalized symmetric FastICA estimator and derive its asymptotic covariance matrix. A review of the literature is also given. We show that the CRB is still achievable.
The concluding remarks of Section VI bring the paper to an end.

\section{ICA data model and method}
In what follows, we denote scalars by lowercase letters $(a,b,c,\ldots)$, vectors by
boldface lowercase letters $(\a,\b,\cc, \ldots)$, and matrices by
 boldface uppercase letters such as $(\A,\B,\C,\ldots)$.  Besides, Greek letters $(\alpha,\beta,\gamma,\ldots)$ usually stand for scalar quantities that play an important role in this work.
 We denote by $\A^\TR$ the matrix transpose of $\A$ and by $\|\A\|$ its spectral norm. With some abuse of notation,  $\|\cdot\|$ also stands for the Euclidean norm for vectors.

\subsection{ICA Data model with infinite sample size\label{sectionIIA}}
We consider the following noiseless linear ICA model:
\ben
\y=\H\s,\label{ICAmodel1}
\een
where
\begin{enumerate}
\item $\s\defeq(s_1,\ldots, s_d)^\TR$ denotes the unknown \emph{source signal}. The components $s_1,\ldots, s_d$ are statistically independent and at most one of them is Gaussian.
\item $\y\defeq(y_1,\ldots, y_d)^\TR$ denotes the \emph{observed signal}.
\item $\H$ is a full rank square matrix, called the mixing matrix.
\end{enumerate}
In ICA model (\ref{ICAmodel1}), the source signal $\s$ and the mixing matrix $\H$ are unknown, while only $\y$ is observable. When the sample size is infinite,
   the probability distribution of $\y$ can be perfectly inferred from the observation, and we can therefore evaluate the mathematical expectation $\Eb[f(\y)]$ for any measurable function $f$. An ICA with the assumption of an  infinite sample size shall hereafter be referred to as the \emph{theoretical ICA}.

The task of ICA is to recover the source signal $\s$ based on the observation $\y$ only. This can apparently be achieved by estimating the inverse of the mixing matrix $\H$. Note that since neither $\H$ nor $\s$ is known, we cannot determine the variance of $\s$. This indeterminacy can be eliminated by fixing {a priori} the variance of $\s$. In this paper, we make the popular convention $\cov(\s)=\I$.

ICA model (\ref{ICAmodel1}) can be simplified by standardizing the observed signal. This procedure consists of the data centering and data whitening:
 \ben\label{theoreticalPreprocess}
 \x  \defeq \cov(\y)^{-\frac{1}{2}}(\y-\Eb[\y]).
 \een
 The standardized signal $\x$ clearly satisfies $\Eb[\x]=0$ and $\cov(\x)=\I$. It can be thought of as the observed signal of the model
\ben\label{standardizedICA}
\x={\A}{{\z}}
\een
 with $\z=\s-\Eb[\s]$ and
${\A}=\cov(\y)^{-1/2}\H=(\H\H^\TR)^{-1/2}\H$. It is easy to see that in the new model the mixing matrix  $\A$ is orthogonal.
Clearly, one can recover $\z$ by estimating $\A^{-1}=\A^\TR$. Due to the inherent ambiguity of ICA \cite{COMO94}, matrix $\A^{-1}$ is only identifiable up the signs and the order of its rows.
 \begin{defi}\label{definition1}
 If a matrix $\W^*$ can be decomposed as $\W^*=\D\P\A^\TR$, where $\P$ is a permutation matrix and $\D$ is a diagonal matrix verifying
 $\D^2=\I$, then $\W^*$ is called a demixing matrix.
\end{defi}
 The ICA then consists of the searching of the demixing matrices on
 the set of orthogonal matrices, that is,
 the orthogonal group $\Ob(d)$.

In the sequel, we call rows of $\W^*$  the \emph{demixing vectors}.
 Denote $\A=(\a_1,\ldots,\a_d)$. Clearly, a vector $\w^*$ can be a demixing vector if and only if there exists some $i\in\{1,\ldots,d\}$ such that $\w^*=\a_i$ or $-\a_i$. Note that when referring to the demixing matrices or demixing vectors, we should keep in mind that the underlying ICA model is the standardized model (\ref{standardizedICA}) rather than the original one (\ref{ICAmodel1}).

\subsection{ICA model with finite sample size}
In practice, we have only a finite sample of $\y$:
 \ben
\y(t)=\H\s(t),\quad t=1,\ldots N\label{ICAmodel3},
\een
where $\y(1),\ldots, \y(N)$ are i.i.d. realizations.
In this case, the standardization procedure
 (\ref{theoreticalPreprocess}) can only be carried out empirically. 
 Natural estimators of $\Eb[\y]$ and $\cov(\y)$ are respectively the empirical mean and empirical covariance matrix:
 \ben
 \bar{\y}&=&\frac{1}{N}\sum_{t=1}^N\y(t),  \nonumber \\
 \hatc&=&\frac{1}{N}\sum_{t=1}^N(\y(t)-\bar{\y})(\y(t)-\bar{\y})^\TR. \label{hatC}
 \een
The empirically standardized data can then be defined as
\ben\label{equation72a}
\x(t)\defeq\hatc^{-1/2}(\y(t)-\bar{\y})=\hata\z(t),
\een
where $\z(t)=\s(t)-\bar{\s}$ and $\hata=\hatc\H$.

Note that care must be taken when dealing with $\x(t)$ and $\z(t)$. Due to the empirical data standardization procedure, neither $\x(1),\dots, \x(N)$ nor $\z(1),\dots, \z(N)$ are independent sequence of random variables.

\section{Variants of the FastICA Algorithm}
\subsection{One-unit FastICA algorithm}
The one-unit version of FastICA, also known as the deflationary FastICA, is the basic form of the algorithm. It searches the local optimizers of the contrast function having the following form:
\ben\label{contrastFunction}\label{contrast1}
\Jc_{1U}(\w)=\Eb[G(\w^\TR\x)],\quad \w\in\Sc,
\een
where $\x$ is the standardized observed signal defined in (\ref{theoreticalPreprocess}) and $G:\Rb\to\Rb$ is a smooth function called the \emph{nonlinearity} or \emph{nonlinearity function}.
In order to be consistent with the notation used in \cite{HYVAOJA2000, HYVA99}, we write $g\defeq G'$, the derivative of $G$. When there is no risk of confusion, both $G$ and its derivative $g$ may be referred to as the ``nonlinearity function''.
Popular nonlinearity  functions \cite{HYVA99} include the following: ``kurtosis'': $x^4/4$, ``gauss'': $-\exp(-\frac{x^2}{2})$ and ``tanh'': $\log\cosh(x)$.

In the sequel, let us denote by $\|\cdot\|$ the $L_2$ norm for vectors and spectral norm for matrices.
The one-unit FastICA algorithm consists of the following steps \cite{HYVA99}:
\begin{enumerate}
\item[1).] Choose  an arbitrary  initial iterate $\w\in\Sc$;
\item[2).] Run iteration
\ben
\w^+ & \leftarrow & \Eb[g'(\w^{\TR}\x)\w - g(\w^{\TR}\x)\x] \label{52} \\
\w & \leftarrow & \frac{\w^+}{\|\w^+\|} \label{52a}
\een
 until convergence.
\end{enumerate}
If one needs to extract more than one source, then an additional orthogonal constraint need to be added between (\ref{52}) and (\ref{52a}):
\ben\label{52b}
\w^+=\w^+-\sum_{i=1}^p\a_i\a_i^\TR\w^+,
\een
where $\a_1,\ldots,\a_p$ are previously obtained demixing vectors. Step (\ref{52b}) is called the deflationary procedure.

Concerning the one-unit FastICA algorithm, we have the following well-known result \cite{HYVAOJA98,HYVA99}:
\begin{prop}\label{hyva}
Let $\w^*$ be a demixing vector corresponding to the extraction of $s_i$.  If
\ben\label{0805a}
\Eb[g'(\w^{*\TR}\x) - g(\w^{*\TR}\x)\w^{*\TR}\x] \\
=\Eb[g'(z_i) - g(z_i)z_i]\neq 0, \nonumber
\een
where $z_i=s_i-\Eb[s_i]$, then
\begin{enumerate}
\item[(i)] $\w^*$ is a fixed point\footnote{We clarify that $\w^*$ is a fixed point in the traditional sense only if the quantity (\ref{0805a}) is strictly positive. If (\ref{0805a}) is negative, then it is well known that the algorithm flips between $\w^*$ and $-\w^*$. In this paper, the latter phenomenon does not cause any problem to the theoretical analysis.
}  of the one-unit FastICA algorithm.
\item[(ii)] It is a local minimizer of $\Jc_{1U}$ on the unit sphere $\Sc$ if $\Eb[g'(z_i) - g(z_i)z_i]> 0$ and local maximizer
if $\Eb[g'(z_i) - g(z_i)z_i]< 0$.
\end{enumerate}
\end{prop}

\subsection{Symmetric FastICA algorithm}
The symmetric version of FastICA extracts all the sources simultaneously. Specifically, it consists of parallel implementations of (\ref{52}) with orthogonal input initial iterates:
\ben
\w_1^+ & \leftarrow & \Eb[g'(\w_1^{\TR}\x)\w_1 - g(\w_1^{\TR}\x)\x]  \label{127c1} \\
 & \vdots &  \nonumber \\
\w_d^+ & \leftarrow & \Eb[g'(\w_d^{\TR}\x)\w_1 - g(\w_1^{\TR}\x)\x],  \label{127c2}
\een
where $\w_1,\ldots,\w_d$ is an orthonormal set. It is then followed by a \emph{symmetrical orthogonalization}, see (\ref{127b}) below.

In the sequel, for any vector $\cc=(c_1,\ldots, c_d)$, we denote
 \be
 \diag(\cc)=\begin{pmatrix} c_1 &&0 \\ & \ddots & \\ 0&&c_d   \end{pmatrix}.
 \ee
 Using matrix notation, we can describe formally the symmetric FastICA algorithm as follows:
\begin{enumerate}
\item[1).] Choose  an arbitrary  orthogonal matrix $\W$;
\item[2).] Run iteration
\ben\label{127a}
\W^+   & \leftarrow & \Eb \Big[\mathrm{diag}\Big(g'(\W\x)\Big)\W - g(\W\x)\x^\TR\Big], \\
\W     & \leftarrow  &    \Big(\W^+\W^{+\TR}  \Big)^{-1/2}\W^+ \label{127b}
\een
 until convergence.
\end{enumerate}

\subsection{Generalized symmetric FastICA algorithm}
The generalized symmetric FastICA algorithm is the same as the ordinary symmetric FastICA, except that it allows the nonlinearity functions used in (\ref{127c1})-(\ref{127c2}) to be different.
In what follows, we denote
\be
\gg(\cc)\defeq\begin{pmatrix} g_1(c_1) \\ \vdots \\ g_d(c_d)  \end{pmatrix},\quad
\gg'(\cc)\defeq\begin{pmatrix} g_1'(c_1) \\ \vdots \\ g_d'(c_d)  \end{pmatrix},
\ee
where $g_i=G_i'$ for $i=1,\ldots,d$ are possibly different nonlinearity functions.
The generalized symmetric FastICA algorithm is defined as follows:
\begin{enumerate}
\item[1).] Choose  an arbitrary  orthogonal matrix $\W$;
\item[2).] Run iteration
\ben\label{127c}
\W^+   & \leftarrow & \Eb \Big[\mathrm{diag}\Big(\gg'(\W\x)\Big)\W - \gg(\W\x)\x^\TR\Big], \\
\W     & \leftarrow  &    \Big(\W^+\W^{+\TR}  \Big)^{-1/2}\W^+
\een
 until convergence.
\end{enumerate}
In this paper, the term ``FastICA algorithm'' always stands for the generalized version of the algorithm unless otherwise specified.

For notational ease, we introduce the following notations:
\ben
\Hc(\W)&\defeq&\Eb \Big[\diag\Big(\gg'(\W\x)\Big)\W - \gg(\W\x)\x^\TR\Big], \label{hc}\\
\Fc(\W)&\defeq&\Big(\Hc(\W) \Hc(\W)^\TR  \Big)^{-1/2}\Hc(\W). \label{fc}
\een
Then the generalized symmetric FastICA algorithm with infinite sample size consists of iterating
\ben \label{equation628b}
\W\leftarrow \Fc(\W)
\een
until convergence.

\subsection{FastICA with Finite sample size}
We have introduced several versions of the FastICA algorithm based on \emph{mathematical expectations}, e.g. (\ref{52}) (\ref{127a}) and (\ref{127c}). The evaluation of mathematical expectation requires a sample of infinite size, which exists only in the theoretical analysis. For this reason,
we shall hereafter refer to the mathematical expectation based FastICA as the \emph{theoretical FastICA}.

In the practical situation, only a sample of finite size is available. Therefore we need to work with an empirical version of those algorithms, which are obtained by approximating the mathematical expectations with the sample means.

For any function $f$, let us denote
\be
\EE[f(\x)]\defeq \frac{1}{N}\sum_{i=1}^N f\big(\x(t)\big).
\ee
Then the empirical version of $\Hc(\cdot)$ and $\Fc(\cdot)$ are given by
\ben
\widehat{\Hc}(\W)&\defeq&\EE \Big[\diag\Big(\gg'(\W\x)\Big)\W - \gg(\W\x)\x^\TR\Big], \label{hathc} \\
\widehat{\Fc}(\W)&\defeq&\Big(\widehat{\Hc}(\W) \widehat{\Hc}(\W)^\TR  \Big)^{-1/2}\widehat{\Hc}(\W). \label{hatfc}
\een
The generalized symmetric FastICA algorithm with finite sample size consists of iterating
\ben\label{equation628c}
\W\leftarrow \hatfc(\W)
\een
until convergence.
We shall refer to algorithm (\ref{equation628c}) as the \emph{empirical FastICA}.

\section{Fixed points of the generalized symmetrical FastICA algorithm }
\subsection{Assumptions\label{assumption}}
This contribution is based on the following regularity conditions:
\begin{enumerate}
\setlength{\itemsep}{1pt}
  \setlength{\parskip}{0pt}
  \setlength{\parsep}{0pt}
\item  For each $i$, the nonlinearity $G_i$ is either even or odd.

\item  The nonlinearities $G_i$ and their derivatives up to the fourth order have polynomial growth:
$|G_{i}^{(k)}(t)| \leq c(|t|^p+1) $ for $i=1,\ldots,d$ and $k=0,1,\ldots,4$, where $c$ and $p$ are some positive constants.
\item  Random vector $\y$ has finite moments until $2p$th order: $\Eb[\|\y\|^{2p}]<\infty$.
\end{enumerate}
These conditions can certainly be weakened, but even in their current form they are convenient to verify and not very restrictive. In fact, it is easy to see that the popular nonlinearities ``kurtosis'', ``gauss'' and ``tanh'' are all even and have polynomial growth with $p=4$. Besides, most common probability distributions have  finite moment of eighth order. Note that we require $G_i$ to be either even or odd
so that a vector $\v$ is a fixed point of one-unit FastICA (\ref{52}) if and only if $-\v$ is also a fixed point. This property will be exploited in the convergence analysis of the algorithm
 (see e.g. Appendix \ref{201561a}). Without any loss of generality, we shall hereafter assume that
$G_i$ are all even for simplicity.
 As for Assumption (2) and assumption (3), these are merely regularity conditions made in order  to satisfy the requirement for
the Uniform Strong Law of Large Numbers (USLLN, Appendix \ref{201561b}) and that for the method of M-estimator (see Appendix \ref{proof_main2}).


\subsection{Characterization of fixed points}
For demixing matrix $\W^*=\D\P\A^{\TR}$, we denote by $\sigma$ the permutation
over $\{1,\ldots,d\}$ induced by $\P$:
\be
\P\z=(z_{\sigma(1)},\ldots,z_{\sigma(d)})^\TR.
\ee
Define
\ben\label{equation625a}  \label{alphai}
\alpha_{i}\defeq \Eb[g_i'(z_{\sigma(i)}) - g_i(z_{\sigma(i)})z_{\sigma(i)}],
\een
which depends on the nonlinearity $g_i$ and the permutation $\sigma$.
We point out that when implementing the generalized FastICA algorithm with different nonlinearities, \emph{a priori} we do not known which nonlinearity is assigned to extract which source.
This issue will be addressed in Section \ref{Cramer}.
Nevertheless, in most part of the work the permutation $\sigma$ does not play an important role and the readers may think of $\sigma$ as the identity permutation when it is not specified in the context.


Now we give a proper definition for the fixed points of the generalized symmetric FastICA algorithm.
 \begin{defi}\label{definition627a}
A matrix $\W$ is defined to be a fixed point of theoretical FastICA (resp. empirical FastICA) if  $\Fc(\W)=\Lbd\W$ (resp. $\hatfc(\W)=\Lbd\W$),
where $\Lbd$ is some diagonal matrix such that $\Lbd^2=\I$.
\end{defi}
It is easily seen that if $\Fc(\W)=\Lbd\W$, then $\Fc(\Lbd\W)=\W$.
 We do not require $\Fc(\W)=\W$ in the definition due to the well-known flipping-sign phenomenon.
 In fact, in most cases such $\W$ does not exist.

\begin{figure*}[t]
\centerline{
\includegraphics[width=1.2\textwidth]{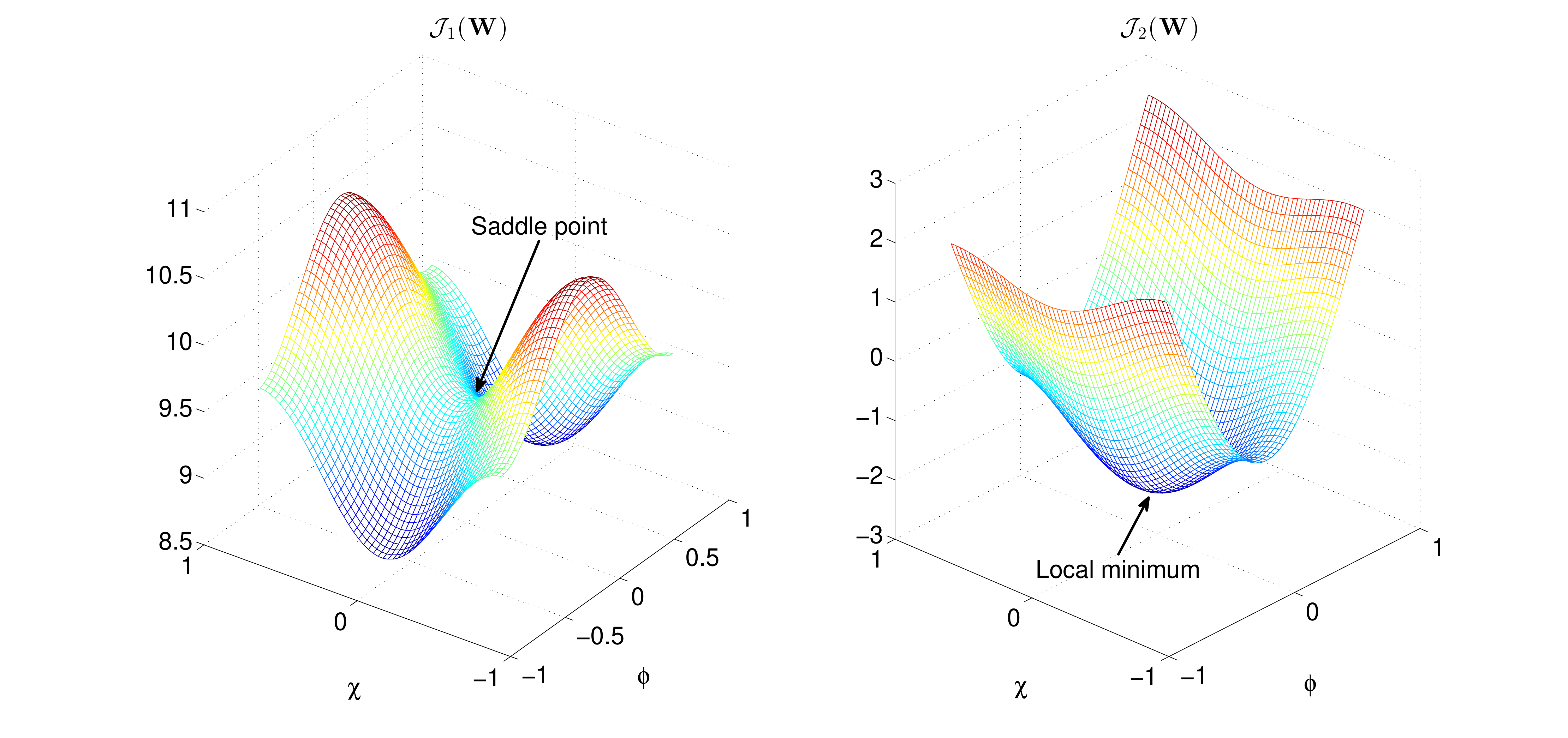}}
\caption{The identity matrix $\I$ at $(\phi,\chi)=(0,0)$ is a saddle point of $\Jc_1$ and is a local minimizer of $\Jc_2$.\label{localcontrast}}
\end{figure*}

The main result of this section is the following theorem:
 \begin{theo}\label{theorem626a}\label{theorem1}
Given a demixing matrix $\W^*$, we suppose that $\alpha_i\neq 0$ for all $i$.
\begin{enumerate}
\item[(i)] Demixing matrix $\W^*$ is a fixed point of the theoretical FastICA algorithm. It is also a local minimizer of
 \ben\label{equation625e}\label{local1}
{\Jc}_{\W^*}(\W)\defeq\sum_{i=1}^d \sign(\alpha_{i})\Eb[G_i(\w_i^\TR\x)]
\een
on the orthogonal group $\Ob(d)$.
\item[(ii)] The empirical FastICA algorithm has almost surely a fixed point $\hatw$ in a neighbourhood of $\W^*$,
which is also a local minimizer of
\ben
\widehat{\Jc}_{\W^*}(\W)\defeq \sum_{i=1}^d \sign(\alpha_{i})\EE[G_i({\w}_i^\TR{\x})] \label{518a}\label{local2}
\een
on $\Ob(d)$ for large enough $N$.
\end{enumerate}
\end{theo}
\begin{IEEEproof} We give the proof for statement (i) here. The proof for statement (ii) can be found in Appendix \ref{proofii}.

We begin by showing that $\W^*$ is a fixed point of $\Fc$. First, it is easy to see that for demixing vector $\w^*_i$, the FastICA update (\ref{52}) yields $\alpha_i\w^*_i$. Since the mapping $\Hc$ is merely the parallel implementations of $(\ref{52})$ with different nonlinearity functions, we get immediately
$\Hc(\W^*)=\L\W^*$, where $\L=\diag(\alpha_1,\dots,\alpha_d)$. It then follows from (\ref{fc}) and the fact $\W^*\in\Ob(d)$ that
\be
\Fc(\W^*)&=&\Big((\L\W^*) (\L\W^*)^\TR  \Big)^{-1/2}(\L\W^*)= \Lbd\W^*,
\ee
where $\Lbd=\diag\big(\sign(\alpha_1),\ldots,\sign(\alpha_d)\big)$.
This means $\W^*$ is indeed a fixed point of $\Fc$ in view of Definition \ref{definition627a}.

Next, we show that $\W^*$ is a local minimizer of (\ref{local1}) on $\Ob(d)$.
By Proposition \ref{hyva}, demixing vector $\w^*_i$ is a local minimizer of $\Eb[G_i(\w^\TR\x)]$ on the unit sphere $\Sc$ if $\alpha_i>0$, and it is a local maximizer of $\Eb[G_i(\w^\TR\x)]$ if $\alpha_i<0$. It follows that
$\w^*_i$ is a local minimizer of $\sign(\alpha_i)\Eb[G_i(\w^\TR\x)]$ for \emph{all} $i=1,\ldots, d$.
This fact suggests that $\W^*$ is a local minimizer of (\ref{local1}) on
\be
\Omega=\{\W\in\Rb^d: \w_i\in\Sc,\quad i=1,\ldots, d\}.
\ee
Since $\Ob(d)\subset \Omega$, matrix $\W^*$ is a local minimizer of (\ref{local1}) on $\Ob(d)$ as well.
\end{IEEEproof}
\begin{rema}
It might be worth pointing out that minimizing the contrast function (\ref{local1}) in the case of ``kurtosis'' nonlinearity
$g_i(s)= s^3$ for all $i=1,\ldots,d$, is equivalent to
 maximizing $\Jc(\W)= \sum_{i=1}^d | \Eb[ (\w^T_i \x)^4] - 3|$
  which is the classical approach  as in Hyv\"{a}rinen \cite{HYVA99}.
\end{rema}

\subsection{Local contrast function}
Theorem \ref{theorem1} reveals the underlying ``contrast function'' of the generalized symmetric FastICA algorithm.
Since this function only has a sense in a neighborhood of a demixing matrix, we are tempted to refer to (\ref{local1}) and (\ref{local2}) as the \emph{local contrast function at }$\W^*$.

 It is interesting to see that the local contrast function ${\Jc}_{\W^*}$ defined in (\ref{local1}) is a sum of the classical contrast functions (\ref{contrast1}) used in the context of one-unit FastICA, with a \emph{correction of sign}, i.e. $\sign(\alpha_{i})$.
 Intuitively, this coefficient $\sign(\alpha_{i})$ serves to make sure that the $i$th column $\w^*_i$ of the demixing matrix $\W^*$ is a local minimizer of $\sign(\alpha_{i})\Eb[G_i(\w^\TR_i\x)]$ on $\Sc$
simultaneously for \emph{all} $i$, so that $\W^*$ can be a local minimizer of
 $\Jc_{\W^*}(\W)$ 
  on $\Ob(d)$.
 If $\sign(\alpha_{i})$ was removed from (\ref{local1}), then $\W^*$ could be a saddle point. See Fig. \ref{localcontrast} and Example \ref{0813a} below.
\begin{exam}\label{0813a}
The purpose of this example is to show that the correction of sign appeared in the local contrast function
(\ref{local1}) cannot be omitted.
Consider the case $d=3$, $G_i=$``kurtosis'' for $i=1,2,3$, $s_1,s_2\sim Uniform$ and $s_3\sim Laplace$.
For simplicity we take the identity mixing matrix $\I$. Denote
\be
\Jc_1(\W) &\defeq& \Eb[G(\w_1^\TR\x)] + \Eb[G(\w_2^\TR\x)] + \Eb[G(\w_3^\TR\x)], \\
\Jc_2(\W) 
&\defeq&\Eb[G(\w_1^\TR\x)] + \Eb[G(\w_2^\TR\x)] - \Eb[G(\w_3^\TR\x)].
\ee
Function $\Jc_1$ is the sum of $\Eb[G(\w_i^\TR\x)]$ without the correction of sign; function $\Jc_2$ is the local contrast function at $\W^*=\I$.
 In fact, we have
 \be
  \alpha_1&=&\alpha_2=\Eb[g'(s_1)-g(s_1)s_1]=1.2>0.\\
  \alpha_3&=&\Eb[g'(s_3)-g(s_3)s_3]=-3.0<0.
  \ee
Therefore, according to (\ref{local1}), only the sign of the term $\Eb[G(\w_3^\TR\x)]$ needs to be altered.

We would like to inspect the values of $\Jc_1(\W)$ and $\Jc_2(\W)$ in a neighborhood of $\I$. Consider the following parametrization:
\be
\W(\phi,\chi)=\begin{pmatrix}
\cos(\phi) & -\sin(\phi)\cos(\chi) & \sin(\phi)\sin(\chi) \\
\sin(\phi) & \cos(\phi)\cos(\chi)  & -\cos(\phi)\sin(\chi)\\
0          & \sin(\chi)            & \cos(\chi)
\end{pmatrix}.
\ee
The set $\{\W(\phi,\chi): \phi,\chi\in[-\pi, \pi)\}$ is a subset of $\Ob(3)$. It
has only two degrees of freedom thus suitable for the 3D plot. We have plotted
the values of $\Jc_1$ and $\Jc_2$ versus the pair $(\phi, \chi)$ and the result is given in Fig. \ref{localcontrast}.
It is easy to see from the figure that
the identity matrix $\I=\W(0,0)$ is a saddle point of $\Jc_1$ but a local minimizer of
$\Jc_2$.
\end{exam}

\begin{exam}\label{0805b}
The purpose of this example is to show that local contrast functions are literally local, in the sense that at different demixing matrices, these functions may be different.
Let us consider a 3-dimensional example with different nonlinearities $G_1=gauss$, $G_2=tanh$, $G_3=kurtosis$ and different sources
  $s_1\sim Laplace$, $s_2\sim GG(4)$ and $s_3\sim Uniform$.
 Here $GG(4)$ stands for the generalized Gaussian distribution with parameter $\alpha=4$, see Appendix
 \ref{GG} for more details.
  Assume an identity mixing matrix $\H=\I$, so that the demixing matrices have the simple form $(\e_{\sigma(1)}, \e_{\sigma(2)}, \e_{\sigma(3)})$, where $\sigma$ is any permutation of $\{1,2,3\}$ and $\e_i$ is the $i$th column of the $3\times 3$ identity matrix $\I$. Here, the sign ambiguity is omitted for simplicity.

   For a demixing matrix
 $\W^*_1=\I=(\e_1,\e_2,\e_3)^\TR$, the associated permutation $\sigma_1$ is the identity permutation. In this case, we have according to (\ref{equation625a})
\be
\alpha_1&=& \Eb[g_1'(z_{1}) - g_1(z_{1})z_{1}]=0.211 >0 \\
\alpha_2&=& \Eb[g_2'(z_{2}) - g_2(z_{2})z_{2}]=-0.077 <0\\
\alpha_3&=& \Eb[g_3'(z_{3}) - g_3(z_{3})z_{3}]=1.200 >0,
\ee
where $z_i=s_i$ for $i=1,2,3$ since all sources have zero mean. It then follows from (\ref{local1}) that
\be
\Jc_{\W^*_1}(\W)=\Eb[G_1(\w^\TR_1\x)] - \Eb[G_2(\w^\TR_2\x)] + \Eb[G_3(\w^\TR_3\x)].
\ee

Likewise, for another demixing matrix, e.g. $\W^*_2=(\e_3,\e_2,\e_1)^\TR$, the associated permutation $\sigma_2$ is the transposition $\sigma_2(1)=3$ and $\sigma_2(3)=1$. Therefore,
\be
\alpha_1&=& \Eb[g_1'(z_{3}) - g_1(z_{3})z_{3}]=-0.217 <0 \\
\alpha_2&=& \Eb[g_2'(z_{2}) - g_2(z_{2})z_{2}]=-0.077 <0\\
\alpha_3&=& \Eb[g_3'(z_{1}) - g_3(z_{1})z_{1}]=-2.990 <0,
\ee
hence
\be
\Jc_{\W^*_2}(\W)=-\Eb[G_1(\w^\TR_1\x)] - \Eb[G_2(\w^\TR_2\x)] - \Eb[G_3(\w^\TR_3\x)].
\ee
 Table \ref{table1} summarizes the local contrast functions at different demixing matrices. As we can see, some of these functions are identical, others are not.  \end{exam}
\begin{table}
  \caption{An example with $d=3$, $\H=\I$, nonlinearities $G_1=$``gauss'', $G_2=$``tanh'', $G_3=$``kurtosis''; source signals
  $s_1\sim$ Laplace, $s_2\sim$ GG(4) and $s_3\sim$ Uniform. \label{table1}}
\centering
{
\begin{Tabular}[2]{c||c}
\hline
 $\W^*$  &  Corresponding $\Jc_{\W^*}$ \\ \hline
$(\e_1, \e_2, \e_3)^\TR$ & $\,\,\,\, \Eb[G_1(\w^\TR_1\x)] - \Eb[G_2(\w^\TR_2\x)] + \Eb[G_3(\w^\TR_3\x)]$     \\
$(\e_1, \e_3, \e_2)^\TR$ & $\,\,\,\, \Eb[G_1(\w^\TR_1\x)] - \Eb[G_2(\w^\TR_2\x)] + \Eb[G_3(\w^\TR_3\x)]$     \\
$(\e_2, \e_1, \e_3)^\TR$ & $-\Eb[G_1(\w^\TR_1\x)] + \Eb[G_2(\w^\TR_2\x)] + \Eb[G_3(\w^\TR_3\x)]$     \\
$(\e_2, \e_3, \e_1)^\TR$ & $-\Eb[G_1(\w^\TR_1\x)] - \Eb[G_2(\w^\TR_2\x)] - \Eb[G_3(\w^\TR_3\x)]$     \\
$(\e_3, \e_1, \e_2)^\TR$ & $-\Eb[G_1(\w^\TR_1\x)] + \Eb[G_2(\w^\TR_2\x)] + \Eb[G_3(\w^\TR_3\x)]$     \\
$(\e_3, \e_2, \e_1)^\TR$ & $-\Eb[G_1(\w^\TR_1\x)] - \Eb[G_2(\w^\TR_2\x)] - \Eb[G_3(\w^\TR_3\x)]$     \\   \hline
\end{Tabular}}
\end{table}

\subsection{On the affine equivariance property \label{equivariance}}
 An ICA method $\Mc(\cdot): \Rb^{d\times N}\to
\Rb^{d\times d}$ that estimates $\H^{-1}$ is called \emph{affine equivariant} if $\Mc(\R\Y) = \Mc(\Y)\R^{-1}$ up to signs and permutation for any full-rank matrix $\R\in\Rb^{d\times d}$, where
$\Y=(\y(1),\ldots,\y(N))$ is the matrix of the observed signals.

Before investigating the affine equivariance property of the generalized symmetric FastICA algorithm, we need to introduce some notations first.
We denote by
$\X=(\x(1),\ldots,\x(N))$ the whitened version of $\Y$.
We write $\Mc(\W,\Y)$ the generalized symmetric FastICA estimator of $\H^{-1}$
 with initial input matrix $\W\in\Ob(d)$ and data matrix $\Y$. We write also
 $\tilde{\Mc}(\W,\X)$ the limit of the generalized symmetric FastICA algorithm
 with initial input matrix $\W$ and the whitened data matrix $\X$. Clearly, the two matrices $\Mc(\W,\Y)$ and $\tilde{\Mc}(\W, \X)$ are related by
 \ben\label{20150705a}
 \Mc(\W,\Y)=\tilde{\Mc}(\W, \X)\widehat{\C}^{-1/2},
 \een
 where $\widehat{\C}$ is the empirical covariance matrix of $\Y$ defined in (\ref{hatC}).

By definition, the generalized symmetric FastICA estimator
with initial input matrix $\W$ is affine equivariant if and only if
\ben\label{20150705b}
 \Mc(\W,\R\Y)= \Mc(\W,\Y)\R^{-1}
\een
up to signs and permutation for any full-rank matrix $\R\in\Rb^{d\times d}$.  Note that the whitened version of $\R\Y$ is
 $\Q\X$, where
 \ben\label{20150705c}
 \Q=(\R\widehat{\C}\R^\TR)^{-1/2}\R\widehat{\C}^{1/2}
 \een
  is an orthogonal matrix.
It then follows from (\ref{20150705a}) that
\ben\label{20150705d}
\Mc(\W,\R\Y) = \tilde{\Mc}(\W, \Q\X)(\R\widehat{\C}\R^\TR)^{-1/2}.
\een
Besides, from (\ref{hathc}) and (\ref{hatfc}) we deduce that
\ben\label{20150705e}
\tilde{\Mc}(\W,\Q\X)=\tilde{\Mc}(\W\Q,\X)\Q^\TR.
\een
Combining (\ref{20150705c})-(\ref{20150705e}) yields
\ben\label{20150705f}
\Mc(\W,\R\Y)=\tilde{\Mc}(\W\Q, \X)\widehat{\C}^{-1/2}\R^{-1}.
\een
Finally, from (\ref{20150705a}) and (\ref{20150705f}) we conclude that (\ref{20150705b}) holds if and only if
\ben\label{20150705g}
\tilde{\Mc}(\W\Q, \X) = \tilde{\Mc}(\W, \X)
\een
up to signs and permutation for any $\Q\in\Ob(d)$. This means that the generalized symmetric FastICA algorithm is affine equivariant if and only if the algorithm is \emph{invariant} with respect to the choice of the initial input matrix.


Based on this characterization, we can now assert that the generalized symmetric FastICA algorithm is in general \emph{not} {affine equivariant}.
This is because different initial input matrices may result in a different assignment of the nonlinearities
$g_1,\ldots,g_d$ to the sources $z_1,\ldots,z_d$, and when it happens,
(\ref{20150705g}) may not hold.


In contrast, the ordinary symmetric FastICA algorithm is affine equivariant, on the condition that there do not exist spurious solutions (this is the case if e.g. the ``kurtosis'' nonlinearity is used \cite{WEI3}).
To prove this, it suffices to show that for arbitrary
initial input matrices $\W_1,\W_2\in\Ob(d)$ we have
$\tilde{\Mc}(\W_1, \X) = \tilde{\Mc}(\W_2, \X)$ up to signs and permutation.
Since there is no spurious solutions, we may assume
 that $\tilde{\Mc}(\W_1, \X)$  is the fixed point in the neighborhood of a demixing matrix $\W^*$ and $\tilde{\Mc}(\W_2, \X)$ the fixed point in the neighborhood of another demixing matrix $\U^*.$
According to Theorem \ref{theorem1}, we have
\be
\tilde{\Mc}(\W_1, \X) &=& \argmin{\W\in\Ob(d)} \widehat{\Jc}_{\W^*}(\W), \\
\tilde{\Mc}(\W_2, \X) &=& \argmin{\W\in\Ob(d)} \widehat{\Jc}_{\U^*}(\W).
\ee
 Since the nonlinearities are all identical,
$\widehat{\Jc}_{\W^*}(\cdot)$ and   $\widehat{\Jc}_{\U^*}(\cdot)$  are related by
\ben\label{0707c}
 \widehat{\Jc}_{\W^*}(\W) =\widehat{\Jc}_{\U^*}(\Q\W),  \quad\forall\,\,\W\in\Ob(d),
\een
  where $\Q=\D\P$ is a matrix that depends on $\W^*$ and $\U^*$,
 here $\P$ is a permutation and $\D$ is diagonal with diagonal entries $\pm 1$.
   It then follows from (\ref{0707c}) that $\tilde{\Mc}(\W_1, \X)$ is a local minimizer of
 $\widehat{\Jc}_{\W^*}$ over $\Ob(d)$ in the neighborhood of $\W^*$  if and only if $\Q\tilde{\Mc}(\W_1, \X)$ is a local minimizer of  $\widehat{\Jc}_{\U^*}$
 over $\Ob(d)$ in the neighborhood of $\U^*$. Then by the unicity of the minimizer, we get $\Q\tilde{\Mc}(\W_1, \X) = \tilde{\Mc}(\W_2, \X)$, which achieves the proof.


Admittedly, the affine equivariance is a desirable property that the generalized FastICA does not enjoy.
Nevertheless, we notice that the method \cite{MIET2014a} originally designed to render the deflation-based FastICA algorithm affine equivariant can also be applied to the generalized symmetric FastICA algorithm. The readers are referred to the paper for more details.

\section{Asymptotic analysis of the generalized symmetric FastICA algorithm\label{sectionV}}
Throughout this section, we shall fix a demixing matrix $\W^*=\D\P\A^{\TR}$ and an initial iterate matrix $\W_0$ that is close enough to $\W^*$. We shall consider the outcome $\widehat{\W}$ of the generalized FastICA algorithm starting at $\W_0$. By Theorem \ref{theorem1}, $\widehat{\W}$ is almost surely near $\W^*$ and shall be considered as an estimator of $\W^*$.

\subsection{Objective of the asymptotic analysis}
\begin{figure*}[!t]
\setcounter{mytempeqncnt}{\value{equation}}
\setcounter{equation}{27}
\ben\label{PSI}
\PSI(\teta,\y)\defeq
\begin{pmatrix}
\y-\muu\\
\w_1^\TR(\y-\muu)(\y-\muu)^\TR\w_1 - \delta_{11} \\
\w_1^\TR(\y-\muu)(\y-\muu)^\TR\w_2 - \delta_{12} \\
\vdots \\
\w_d^\TR(\y-\muu)(\y-\muu)^\TR\w_d - \delta_{dd} \\
\tilde{g}_1\big(\w_1^\TR(\y-\muu)\big)\w_2^\TR(\y-\muu) - \tilde{g}_2\big(\w_2^\TR(\y-\muu)\big)\w_1^\TR(\y-\muu) \\
\tilde{g}_1\big(\w_1^\TR(\y-\muu)\big)\w_3^\TR(\y-\muu) - \tilde{g}_3\big(\w_3^\TR(\y-\muu)\big)\w_1^\TR(\y-\muu) \\
\vdots\\
\tilde{g}_{d-1}\big(\w_{d-1}^\TR(\y-\muu)\big)\w_{d}^\TR(\y-\muu) - \tilde{g}_{d}\big(\w_{d}^\TR(\y-\muu)\big)\w_{d-1}^\TR(\y-\muu)
\end{pmatrix}
=\begin{pmatrix}
\PSI_{\muu}\\
\PSI_{11} \\
\PSI_{12} \\
\vdots \\
\PSI_{dd} \\
\PSI_{\w_1\w_2} \\
\PSI_{\w_1\w_3} \\
\vdots\\
\PSI_{\w_{d-1}\w_{d}}
\end{pmatrix},
\een
\hrulefill
\vspace*{4pt}
\end{figure*}
We recall that in the context of FastICA (and many other ICA methods), we do not directly work with the observed signal $\y(t)$, rather, we work with the centered and whitened data
\be
\x(t)
&\defeq&\hatc^{-1/2}(\y(t)-\bar{\y})\\
&=&\hatc^{-1/2}\H(\s(t)-\bar{\s})=  \hata\z(t),
\ee
where $\hata=\hatc^{-1/2}\H$, $\z(t)=\s(t)-\bar{\s}$
and $\hatc$ is the empirical covariance matrix of $\y$:
\be
\hatc&=&\frac{1}{N}\sum_{t=1}^N(\y(t)-\bar{\y})(\y(t)-\bar{\y})^\TR.
\ee
%
The demixing matrix $\W^*$ naturally defines a solution $\B$ of the original ICA model (\ref{ICAmodel1}), that is, a matrix equal to $\H^{-1}$ up to the signs and a permutation of its rows:
\be
\B\defeq \W^*\cov(\y)^{-1/2}=\D\P\H^{-1}.
\ee
 Similarly, the generalized symmetric FastICA estimator $\widehat{\W}$ of $\W^*$,
 yields an estimator $\widehat{\B}$ of $\B$:
\ben\label{523}
\widehat{\B}\defeq \widehat{\W}\hatc^{-1/2}.
\een

Let $\b^\TR_i$ and $\hat{\b}^\TR_i$ be respectively the $i$th row of $\B$ and $\widehat{\B}$ for $i=1,\ldots,d$.
The objective of this section is to derive for each $i$ the limiting distribution of
${N}^{1/2}(\hat{\b}_i - \b_i)$ and the
asymptotic rate of the convergence $\hat{\b}_i\to\b_i$ as the sample size $N$ tends to infinity.

\subsection{M-estimator}
The method of estimating equation and M-estimator \cite{VAAR} is a powerful tool to solve problems of this kind, see\cite{HYVA2006, OLLI2011, REYH2012, WEI2} for some earlier results based on this method.

Let us suppose that the unknown distribution of random vector $\y$ depends on some parameter $\teta$ of interest. Suppose also that the true parameter is $\teta^*$, which satisfies equation $\Eb[\PSI(\teta^*,\y)]=0$, where $\PSI$ is some vector valued function. Let
$\y(1),\ldots,\y(N)$ be an i.i.d. sample of $\y$. Then an estimator $\hat{\teta}$ is obtained by solving the following equation
\ben\label{219a}
\frac{1}{N}\sum_{t=1}^N\PSI\big({\teta}, \y(t)\big)=\EE[\PSI(\teta,\y)]=0.
\een
The estimator $\hat{\teta}$ is called an M-estimator, and equation (\ref{219a}) is called the estimating equation.
Under some mild regularity conditions (see Appendix \ref{proof_main2}),  there holds
\ben\nonumber
N^{1/2}(\hat{\teta}-\teta^*)\CD \Nc\Big(0,   \Q^{-1}\Eb[\PSI(\teta^*,\y) \PSI(\teta^*,\y)^\TR]\Q^{-\TR}\Big),\\
\label{0806c}
\een
where $\CD$ denotes the convergence in distribution and
\be
\Q\defeq \Eb\Big[\frac{\partial}{\partial{\teta}}\PSI(\teta,\y)\Big|_{\teta=\teta^*}\Big].
\ee

To apply this result, we need to
\begin{enumerate}
\item Find an appropriate function $\PSI(\teta,\y)$ for our problem;
\item Compute the matrix
\ben\label{0806b}
\Q^{-1}\Eb[\PSI(\teta^*,\y) \PSI(\teta^*,\y)^\TR]\Q^{-\TR}.
\een
\end{enumerate}

To achieve step 1), we shall rely on the characterization established in Theorem \ref{theorem1}, which states that  $\W^*$ and $\widehat{\W}$
are solutions of two related constrained optimization problems. This implies that they must satisfies the
Kuhn-Tuker first order necessary conditions \cite{FIOR2006,ERDO}. Based on these, we are able to derive the following lemma:
\begin{lemm}\label{lemma626a}\label{optimum_condition}
Let $\W^*$ be a fixed demixing matrix.
Denote $\tilde{G}_i=\sign(\alpha_i)G_i$ and $\tilde{g}_i=\sign(\alpha_i)g_i$ for $i=1,\ldots,d$.
If a matrix $\U$ is a solution of
\ben\label{0807a}
 \min_{\W\in\Ob(d)}\sum_{i=1}^d \Eb[\tilde{G}_i(\w_i^\TR\x)],
\een
 then
$\Eb[\boldsymbol{\tilde{g}}(\U\x)\x^\TR\U^\TR]$  
is a symmetric matrix.
\end{lemm}
A similar result holds if we replace $\Eb$ by its sample average counterpart $\EE$.


The main difficulty to derive $\PSI(\cdot, \cdot)$ is that it must take the original signal $\y$ as its second argument rather than the standardized signal $\x$. This is because $\y(1),\ldots,\y(N)$ are i.i.d. random variables (for which the method of M-estimator is applicable), while $\x(1),\ldots, \x(N)$ are not, due to the dependency introduced by the data standardization procedure.
We resolve this issue by applying the change of variable $\hat{\muu}=\bar{\y}$, see Appendix \ref{proof_estimating} for more details.
\begin{lemm}\label{estimating_equation}
Let us define mapping
\be
\PSI(\teta,\y): \Rb^{d^2+d}\times \Rb^d\to\Rb^{d^2+d},
\ee
its explicit form being given in (\ref{PSI}), where $\teta\defeq (\W, \muu)$ and $\delta_{ij}$ is the Dirac delta function.
 Then
\begin{enumerate}
\item[(i)] $\hat{\teta}\defeq(\hatb,\hat{\muu})$ is a solution of $\EE[\PSI(\teta, \y)]=0$;
\item[(ii)] ${\teta}^*\defeq(\B,\Eb[\y])$ is a solution of $\Eb[\PSI(\teta, \y)]=0$.
\end{enumerate}
\end{lemm}
\begin{IEEEproof}
See Appendix \ref{proof_estimating}.
\end{IEEEproof}
Next, we need to compute the asymptotic covariance matrix (\ref{0806b}).
The challenge is that $\Q$ being a fairly large
$\Rb^{d^2+d} \times \Rb^{d^2+d}$ matrix, its direct inversion
 is generally a difficult task.
 Fortunately, we do not really need to calculate $\Q^{-1}$ in order to compute the asymptotic covariance matrix.
 In fact, in view of Lemma \ref{estimating_equation}, only the component $\W$ contained in the parameter $\teta$ is interesting to us, while (\ref{0806b}) is the covariance matrix for the entire parameter $\teta=(\W, \muu)$.
 To resolve this problem,
 let us rewrite (\ref{0806c}) as
 \be
&N^{1/2}(\hat{\teta}-\teta^*)\CD  \T,&\\
&\T\sim \Nc\Big(0,   \Q^{-1}\Eb[\PSI(\teta^*,\y) \PSI(\teta^*,\y)^\TR]\Q^{-\TR}\Big)&.
\ee
Clearly, we have
\ben\label{equation2}
\M\defeq\Q\T \sim \Nc\Big(0,   \Eb[\PSI(\teta^*,\y) \PSI(\teta^*,\y)^\TR]\Big).
\een
The idea is to solve (\ref{equation2}) without calculating $\Q^{-1}$ for the component of $\T$. 
Once we obtain $\t_{\w_i}=\K\M$
for some matrix $\K$,
then the covariance matrix of $\t_{\w_i}$ follows.

\subsection{Main result}
Now we are ready to announce the main result of this work:
\begin{theo}\label{main2}
Assume that the  following mathematical expectations exist for $i=1,\ldots,d$:
\ben
\alpha_{i}&\defeq& \Eb[g_i'(z_{\sigma(i)}) - g_i(z_{\sigma(i)})z_{\sigma(i)}] \nonumber \\
\beta_{i} & \defeq & \Eb[g_i(z_{\sigma(i)})^2]  \nonumber  \\
\gamma_{i} & \defeq & \Eb[g_i(z_{\sigma(i)})z_{\sigma(i)}]  \nonumber \\
\eta_{i} & \defeq & \Eb[g_i(z_{\sigma(i)})] \nonumber \\
\tau_{i} & \defeq & (\Eb[z_{\sigma(i)}^4]-1)/4, \nonumber
\een
where $z_i=s_i-\Eb[s_i]$ for $i=1,\ldots,d$.
Then we have
$N^{1/2}(\hat{\b}_i - \b_i)\CD \Nc(0, \R_i)$,
where
\ben
 \R_i
&&=\sum_{j\neq i}^d \frac{\beta_i - \gamma_i^2 + \beta_j - \gamma_j^2 + \alpha_j^2 -\eta_i^2 - \eta_j^2
 }{(|\alpha_i| + |\alpha_j|)^2}\b_j\b_j^\TR+ \tau_i\b_i\b_i^\TR. \nonumber \\
 \label{ASYM2}
\een
\end{theo}
\begin{IEEEproof}
See Appendix \ref{proof_main2}.
\end{IEEEproof}
\begin{rema}
If $s_{\sigma(i)}$ has symmetric distribution, then the quantity $\eta_{i}$ vanishes since $g_i$ is an odd function. Therefore, when all the source signals have symmetric distribution,  formula (\ref{ASYM2}) is reduced to
\be
 \R_i
&&=\sum_{j\neq i}^d \frac{\beta_i - \gamma_i^2 + \beta_j - \gamma_j^2 + \alpha_j^2}{(|\alpha_i| + |\alpha_j|)^2}\b_j\b_j^\TR+ \tau_i\b_i\b_i^\TR.
\ee
\end{rema}
Now we consider  $\G\defeq\B\H$ and
$\widehat{\G} = \widehat{\B}\H$. The former matrix is equal to identity up to a sign and a permutation; the latter matrix, referred to as the gain matrix by some authors \cite{TICHOJA}, represents through its $(i,j)$th element the relative presence of the $j$th source signal in the estimated $i$th source signal.
A number of performance indices of ICA are proposed based on the gain matrix. We refer the readers to
\cite{ILMO2010a} for more details.

By Theorem \ref{main2}, the asymptotic normality and asymptotic variance of $\widehat{\G}\to\G$ can be easily derived.
 Denote $\H=(\hh_1,\ldots,\hh_d)$. Then we have
\be
N^{1/2}(\widehat{\G}_{ij} - \G_{ij})
&\CD&\Nc(0, \hh_{j}^\TR\R_i\hh_{j}),
\ee
where $\G_{ij}$ and $\widehat{\G}_{ij}$ denote respectively the $(i,j)$ entry of $\G$ and $\widehat{\G}$.
Then using the equalities $\h_{\sigma(i)}^\TR\b_i=\pm1$ and $\h_{\sigma(j)}^\TR\b_i=0$ for $j\neq i$, we obtain the following result:
\begin{coro}\label{corollary315}For $i,j=1,\ldots,d$, there holds
\ben\label{20150707a}
N^{1/2}(\widehat{\G}_{i,\sigma(j)} - \G_{i,\sigma(j)})\CD \Nc(0, V_{i,\sigma(j)}),
\een
 where $V_{i,\sigma(i)}=\tau_i$ and
\ben
V_{i,\sigma(j)}&=&\frac{\beta_i - \gamma_i^2 + \beta_j - \gamma_j^2 + \alpha_j^2 -\eta_i^2 - \eta_j^2
 }{(|\alpha_i| + |\alpha_j|)^2},\,\, j\neq i. \quad\label{gain1}
\een
\end{coro}
\begin{figure}[t]
\centerline{
\includegraphics[width=0.5\textwidth]{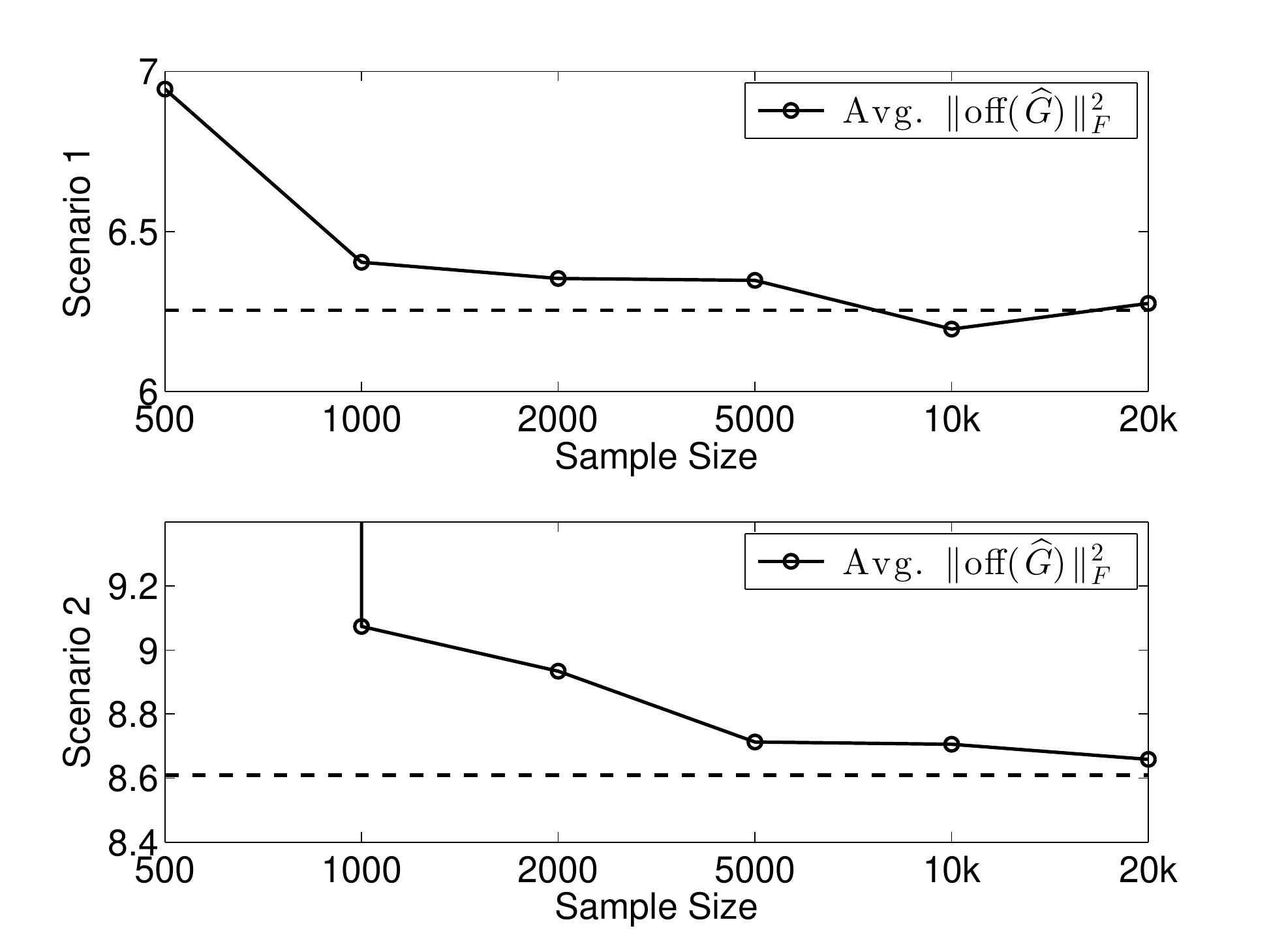}}
\caption{The average of $\|\mathrm{off}(\widehat{\G})\|_F^2$ in 10000 independent trials with different sample sizes. The dashed line indicates
the theoretical mean value of the asymptotic distribution given in (\ref{20150707b}).
In Scenario 1 the algorithm starts at $(\e_1,\e_2,\e_3)^\TR$ while in Scenario 2
it starts at  $(\e_2,\e_3,\e_1)^\TR$.
\label{convergence}}
\end{figure}
\begin{exam}
We intend to validate
(\ref{20150707a}) based on
the \emph{minimum distance index} proposed in \cite{ILMO2010a}.
According to \cite[Theorem 1]{ILMO2010a}, if (\ref{20150707a}) holds then
\ben\label{20150707b}
\lim_{N\to\infty}\Eb[\|\mathrm{off}(\widehat{\G})\|_F^2] = \sum_{i,j=1; i\neq j}^dV_{i,\sigma(j)},
\een
where $\mathrm{off}(\widehat{\G})$ is the matrix obtained by setting the ``diagonal'' entries $\widehat{\G}_{i,\sigma(i)}$ to zero, $i=1,2,3$ and
$\|\cdot\|_F$ denotes the Frobenius norm.
The idea here is to compute the empirical average of $\|\mathrm{off}(\widehat{\G})\|_F^2$ for some large $N$ in many independent trials and compare it with the term on the right-hand side of (\ref{20150707b}).
We consider the case of $d=3$ with three different nonlinearities
$G_1=kurtosis$, $G_2=gauss$, $G_3=tanh$ and three different sources
  $s_1\sim Bimod(3, -0.3)$, $s_2\sim GG(4)$ and $s_3\sim Laplace$.
 Here Bimod($a_1,a_2$) denotes the bimodal Gaussian distribution with two modes at $a_1$ and $a_2$  and $GG(\alpha)$ denotes
 the generalized Gaussian distribution with parameter $\alpha$. More details of the
 distributions used here can be found in Appendix \ref{somePDF}.
 In the simulations,
 the mixing matrix is fixed to be $\H=\I$ and
 a number of different sample sizes, from $N=500$ to $N=20000$, are considered.
Two scenarios are investigated here: In the first scenario, we let
 the generalized symmetric FastICA algorithm start at $\I=(\e_1,\e_2,\e_3)^\TR$,
 so that the nonlinearity $G_i$ is assigned to extract $s_i$ for each $i=1,2,3$.
In the second scenario, the algorithm starts at $(\e_2, \e_3, \e_1)^\TR$, in which case $G_1, G_2, G_3$ are
assigned to extract $s_2, s_3, s_1$ respectively.
 The simulation results are given in Fig. \ref{convergence}, which confirm the validity of (\ref{20150707a}).
 \end{exam}

\subsection{Asymptotic performance with known $\Eb[\y]$}
In this section, we investigate the limiting distribution of the generalized FastICA with the assumption that $\Eb[\y]$ is known. Consider
\be
\widetilde{\C}&\defeq& \frac{1}{N}\sum_{t=1}^N\big(\y(t)-\Eb[\y]\big)\big(\y(t)-\Eb[\y]\big)^\TR,\\
\tilde{\x}(t)&=&\widetilde{\C}^{-1/2}\big(\y(t) - \Eb[\y]\big),\quad t=1,\ldots,N,
\ee
where $\Eb[\y]$ is used instead of $\bar{\y}$.
Denote by $\widetilde{\W}$ the limit of generalized symmetric FastICA
  with input $\tilde{\x}(t)$.
Then $\widetilde{\B}\defeq\widetilde{\W}\widetilde{\C}^{-1/2}$ is an estimator of $\B$.
In this case, the function $\PSI(\cdot,\cdot)$ used in the M-estimator becomes
\ben
\widetilde{\PSI}(\W,\y)\defeq
\begin{pmatrix}
\w_1^\TR\y\y^\TR\w_1 - \delta_{11} \\
\vdots \\
\w_d^\TR\y\y^\TR\w_d - \delta_{dd} \\
\tilde{g}_1\big(\w_1^\TR\y\big)\w_2^\TR\y - \tilde{g}_2\big(\w_2^\TR\y \big)\w_1^\TR\y \\
\vdots\\
\tilde{g}_{d-1}\big(\w_{d-1}^\TR\y\big)\w_{d}^\TR\y- \tilde{g}_{d}\big(\w_{d}^\TR\y\big)\w_{d-1}^\TR\y
\end{pmatrix},\quad\quad
\een
which is clearly different from (\ref{PSI}). Eventually, this leads to
 a different limiting distribution:
 \be
N^{1/2}(\tilde{\b}_i - \b_i)\CD \Nc(0, \widetilde{\R}_i),
\ee
where
\ben
\widetilde{\R}_i &=&\sum_{j\neq i}^d \frac{\beta_i -\gamma_i^2 + \beta_j -\gamma_j^2 + \alpha_j^2 - \eta_j^2}{(|\alpha_i|
 + |\alpha_j|)^2} \b_j\b_j^\TR  + {\tau}\b_i\b_i^\TR  \nonumber   \\
&&+ \sum_{j\neq i}^d \frac{\b_j \eta_j}{(|\alpha_{i}|+|\alpha_{j})|} \sum_{j\neq i}^d \frac{\b_j^\TR \eta_j}{(|\alpha_{i}|+|\alpha_{j}|)} \nonumber       \\
&& -\sum_{j\neq i}^d\frac{ \Eb[s_i^3]\Eb[g(s_j)] }{2(|\alpha_{i}|+|\alpha_{j}|)}  (\b_j\b_i^\TR + \b_i\b_j^\TR).
\label{0808a}\label{ASYM3}
\een
The proof of this result is quite similar to that  of Theorem \ref{main2}. It is omitted here due to the lack of space.
\begin{rema}\label{0808c}
 It is easy to see that the asymptotic variance of the entries of the former gain matrix is given by
\ben
\tilde{V}_{i,\sigma(j)}&=& \h^\TR_{\sigma(j)} \widetilde{\R}_i  \h_{\sigma(j)} \nonumber \\
&=&\frac{\beta_i - \gamma_i^2 + \beta_j - \gamma_j^2 + \alpha_j^2  }{(|\alpha_i| + |\alpha_j|)^2},\,\, j\neq i, \quad\label{gain3}
\een
and $\tilde{V}_{i,\sigma(i)}=\tau_i$. Comparing (\ref{gain3}) with (\ref{gain1}), we observe that $\tilde{V}_{i,\sigma(j)}\geq {V}_{i,\sigma(j)}$ for $j\neq i$ and the equality takes place only if $\eta_i$ and $\eta_j$ vanish. By definition $\eta_i\defeq\Eb[g_i(s_{\sigma(i)})]$, the latter may happen
if both $s_{\sigma(i)}$ and $s_{\sigma(j)}$ have symmetric distributions.
\begin{figure*}[t]
\centerline{
\includegraphics[width=\textwidth]{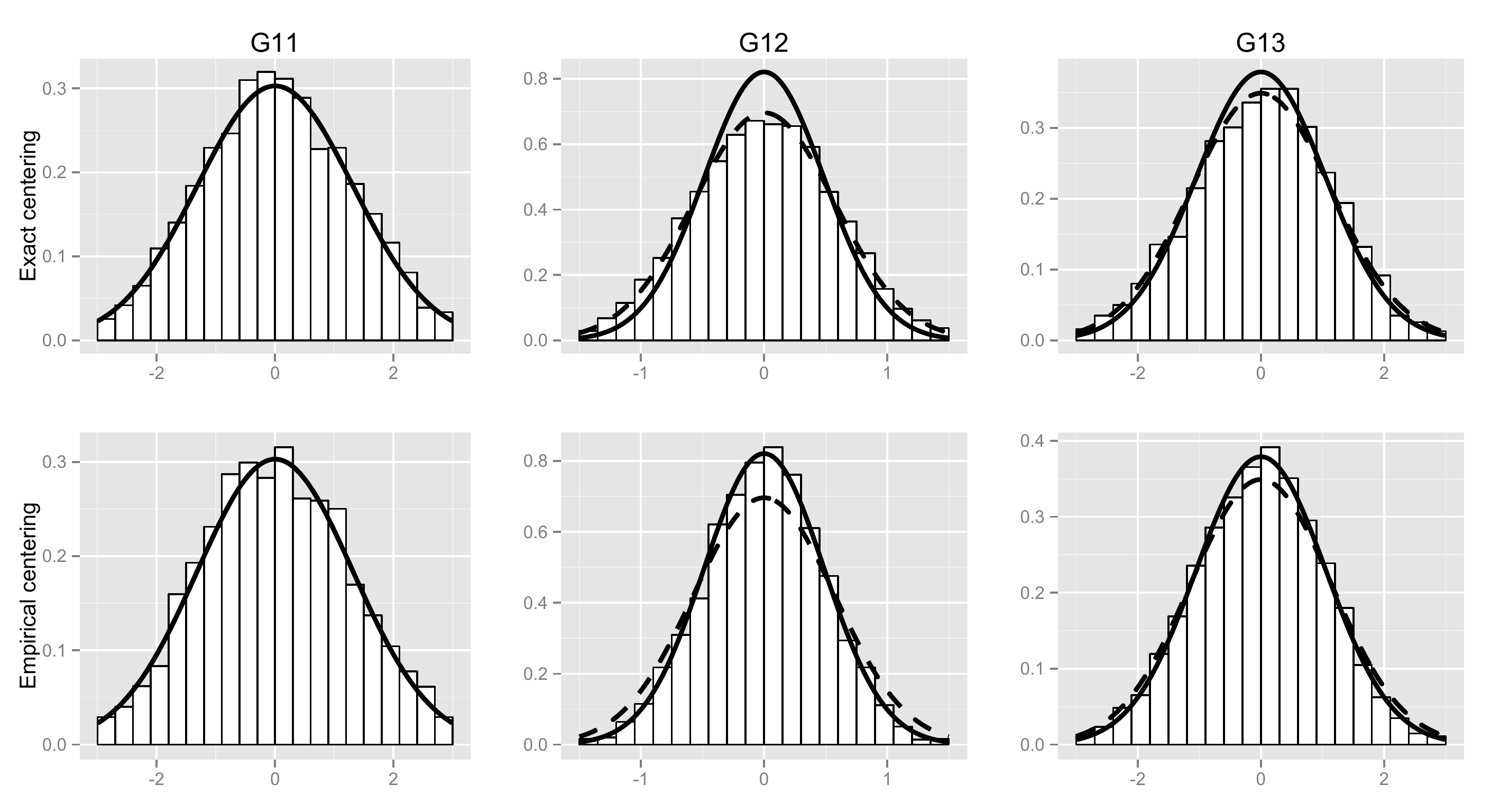}}
\caption{Histograms of $N^{1/2}(\widehat{\G}_{1j} - \G_{1j})$ for $j=1,2,3$ in 5000 independent trials with $N=10000$ versus
the curves of Gaussian PDFs with asymptotic variances given by formula (\ref{gain1}) (in dashed curve) and (\ref{gain3}) (in solid curve). \label{normality3d}}
\end{figure*}
\begin{exam}\label{example1}
In this example we intend to verify the asymptotic normality established in Theorem \ref{main2} as well as the asymptotic variances formulae (\ref{gain1}) and (\ref{gain3}) using histograms. We fix a large sample size $N$ and then compute $N^{1/2}(\widehat{\G}_{ij} - \G_{ij})$ in many independent trials. If the asymptotic normality holds, then the histogram of $N^{1/2}(\widehat{\G}_{ij} - \G_{ij})$ obtained  should resemble that of a normal distribution.
 Here we consider the case $N=10000$, $d=3$ and three different nonlinearity functions: $G_1=$``Gauss'', $G_2=$''Tanh'', $G_3=$``kurtosis''. The sources have identical distribution: $s_i\sim$ Bimod(3,-0.3) for $i=1,2,3$,
 The simulations were carried out in two independent scenarios. In the first scenario, FastICA was implemented with exact data centering (i.e. using $\y(t) - \Eb[\y]$). In the second one, it was implemented with empirical data centering (i.e. using $\y(t)-\bar{\y}$).

 The three figures in the first row corresponds to the case of exact centering.
 The histograms of $N^{1/2}(\widehat{\G}_{1j} - \G_{1j})$ for $j=1,2,3$ and
the curves of Gaussian PDFs with variances given by (\ref{gain1}) (in solid curve) and (\ref{gain3}) (in dash curve) are plotted.
We observe from the first row that the histograms match well the dashed curves in all plots. This observation confirms the validity of (\ref{gain3}). The three figures in the second row corresponds to the scenario of empirical centering. In this case, the histograms match the solid curves, which confirms (\ref{gain1}).
Note that the solid curve and the dashed curve overlap in the two figures of the first column. Therefore
only the solid curve is visible.
\end{exam}


\end{rema}
\subsection{Related work\label{review}}
Many researchers have studied the asymptotic behavior of FastICA \cite{HYVA97,OLLI, OLLI2011, TICHOJA,
 MIET2014a, MIET2014b, VIRT2015}. However, most of the work were dedicated to the one-unit version of the algorithm, which is much easier to deal with.

The first result on this subject seems to be \cite{HYVA97}. In this work,
the author derived the trace of the asymptotic covariance matrix of the one-unit FastICA estimator:
\ben\label{formula_hyva}
\mathrm{Trace}(\R_i)=c\cdot\frac{\beta_i-\gamma_i^2}{\alpha_i^2 },
\een
where $c$ is a constant that depends only on the mixing matrix. The main limitation of the work is that the author
assumed the usage of the exact mean and covariance matrix in the data preprocessing. In other words,
the impact of empirical data centering and whitening was not taken into account.

Paper \cite{OLLI} along with \cite{OLLI2011} tackle the one-unit FastICA with the deflation procedure (\ref{52b}). Using the method of Influence Function, the author  derived a closed-form expression of the asymptotic covariance matrix for the general $i$th sequentially estimated demixing vector using one-unit FastICA:
\ben
\R_i^{1U}& =&  \sum_{j=1}^{i-1}\frac{\beta_j-\gamma_j^2-\eta_j^2+\alpha_j^2}{\alpha_j^2 }\b_j\b_j^\TR + \tau_i\b_i\b_i^\TR \nonumber \\
&&+\frac{\beta_i-\gamma_i^2-\eta_i^2}{\alpha_i^2 } \sum_{j=i+1}^d\b_j\b_j^\TR. \label{formula_olli}
\een
 For the extraction of the first source $s_1$, the expression above is reduced to
\ben
\R_1^{1U}& =&   \frac{\beta_1-\gamma_1^2-\eta_1^2}{\alpha_1^2 } \sum_{j=2}^d\b_j\b_j^\TR + \tau_1\b_1\b_1^\TR. \label{formula_olli2}
\een
Both (\ref{formula_olli}) and (\ref{formula_olli2}) are validated by numerical simulations.

Contribution \cite{MIET2014a} also studies the one-unit FastICA with the deflation procedure (\ref{52b}).
The authors generalized the results of \cite{OLLI} by allowing the usage of different nonlinearities in different deflation stage of one-unit FastICA, in order to achieve a better separation performance. Besides, a method that renders the deflation-based FastICA affine equivariant is also proposed.

Another important work is \cite{TICHOJA}, in which the Cram\'er-Rao lower bounds for ICA, along with asymptotic covariance matrices of both one-unit and symmetric FastICA were derived:
\ben
{V}^{1U}_{i,j}&=& \frac{\beta_i-\gamma_i^2}{\alpha_i^2 }, \,\, j\neq i. \quad\nonumber\\
{V}^{SYM}_{i,j}&=&\frac{\beta_i - \gamma_i^2 + \beta_j - \gamma_j^2 + \alpha_j^2
 }{(|\alpha_i| + |\alpha_j|)^2},\,\, j\neq i. \quad\label{gain2}
\een
Expression (\ref{gain2}) is different from our result (\ref{gain1}) but coincides with (\ref{gain3}).
In our opinion, the main drawback of this result is that it is based on a heuristic approach.
Expression (\ref{gain2}) 
is only valid if all sources involved have symmetrical distributions, as is explained in Remark \ref{0808c}.
In the general case, it is easy to verify that only (\ref{gain1}) is correct, see e.g. Fig \ref{normality3d}.

The most recent work on this subject seems to be \cite{MIET2014b}. The authors derived the limiting distributions for the symmetric FastICA with ``kurtosis'' nonlinearity, along with  several other cumulant-based ICA algorithms.

\subsection{Approaching the Cram\'er-Rao bound \label{Cramer}}
As is explained in Section \ref{equivariance}, the generalized symmetric FastICA algorithm is not affine equivariant by
allowing the usage of different nonlinearities, as the
outcome of the algorithm depends on how the nonlinearities are assigned to the sources. However, there is a gain in separation performance, i.e.
  the algorithm has the potentiel to attain the Cram\'er-Rao lower bound for ICA.

If the source signals have smooth probability density functions (PDF), then their respective \emph{score function} exists:
\be
\psi_i(x)\defeq \frac{f_i'(x)}{f_i(x)},
\ee
where $f_i$ denotes the PDF of the $i$th source signal.
Denote
\be
\kappa_i\defeq \Eb[\psi_i^2(s_i)]=\int_{\Rb}\frac{f_i^{\prime 2}(x)}{f_i(x)}\mathrm{d}x.
\ee
The Cram\'er-Rao bound (CRB) for linear ICA model (\ref{ICAmodel1}) has already been studied in \cite{TICHOJA, OLLI2008}. It is shown that
under some mild conditions, the CRB for the
asymptotic variance of the entry of gain matrix exists and  is equal to
\ben \label{CRB}
\mathrm{CRB}(V_{ij})=\frac{\kappa_i}{\kappa_i\kappa_j-1}.
\een
It is explained in \cite{TICHOJA2} that one can attain the CRB (\ref{CRB}) by choosing the nonlinearities wisely.
The authors proposed to run the generalized symmetric FastICA algorithm multiple times with the optimum nonlinearities for each independent component.
Although this method is based on expression (\ref{gain2}), which does not hold for sources with asymmetric distribution, the conclusion remains valid. In fact, when the optimum nonlinearities are chosen, quantities $\eta_i$ vanishes for all $i$, hence (\ref{gain2}) coincides with (\ref{gain1}).
To see this, let us
fix an index $i$ and take
\ben
g^{\mathrm{opt}}_{i}(x)&=&\psi_i(x) \label{equation78b} \\
g_{j}^{\mathrm{opt}}(x)&=&\frac{1}{\kappa_j}\psi_j(x), \quad\quad j\neq i. \label{equation78a}
\een
Straightforward calculation gives
\be
\beta_i & = & \Eb[\psi_i(s_{i})^2]= {\kappa_i} \\
\gamma_{i} & = & \Eb[\psi_i(s_{i})s_{i}]= 1\\
\alpha_i & = & \Eb[\psi_i'(s_{i}) - \psi_i(s_{i})s_{i}] = {\kappa_i-1}\\
\eta_i &=& \Eb[\psi_i(s_i)]=0.
\ee
while $\beta_j=\gamma_j=\kappa_j^{-1}$, $\alpha_j=1-\kappa_j^{-1}$ and $\eta_j=0$.
Inserting these values in (\ref{gain1}), we obtain
\ben
V_{ij}&=&\frac{\kappa_i-1 + \kappa_j^{-1} - \kappa_j^{-2} + (1-\kappa_j^{-1})^2}{(\kappa_i-1+ 1 - \kappa_j^{-2})^2}=\frac{\kappa_j}{\kappa_i \kappa_j-1} \nonumber \\
&=&\mathrm{CRB}(V_{ij}).  \label{63CRB}
\een
\begin{rema}
There are some practical issues here.
To exploit this result, one needs to 1) estimate $\psi_i(\cdot)$ for each $i$ and 2) assign each
$g_i=\hat{\psi_i}(\cdot)$ to the corresponding $s_i$, while
 \emph{a priori} one knows neither the PDFs nor the direction for each of the sources.
 To resolve these issues, it is proposed \cite{TICHOJA2, MIET2014a} to
obtain first a preliminary demixing matrix via e.g. the ordinary symmetric FastICA with nonlinearity ``tanh'', or another ICA method such as JADE \cite{CARD1993}. Using the preliminary demixing matrix, one can then estimate each score function $\psi_i(\cdot)$ based on the empirical distribution of the extracted source $s_i$, and determine the (approximate) direction for each $s_i$.
Nevertheless, since the  nonlinearity $\hat{\psi}_i$ is only a estimation of $\psi$,
one can only approach the CRB in practice. The readers are referred to \cite{TICHOJA2} for simulation results.
\end{rema}

\section{Conclusion}
The contribution of this work is twofold: 1) It is shown that the algorithm optimizes a function that is a sum of the contrast functions used by traditional one-unit FastICA with a correction of sign; 2) The
limiting distribution of the generalized symmetric FastICA algorithm is derived, and
an original closed-form expression of the asymptotic covariance matrix is given.
Numerical simulations match very well our theoretical prediction.

\appendices
\section{Proof of Theorem \ref{theorem626a} (ii) \label{proofii}}

\subsection{Some preliminary results \label{201561b}}
We will need the Uniform Strong Law of Large Numbers (USLLN). The following version of USLLN can be found in \cite{BIER}.
For a detailed discussion of this theorem, we refer to \cite{NEWE1991, ANDR1992}.
\begin{theo}[USLLN]\label{theoremUSLLN}\label{USLLN1}
Let $\y(1),\ldots, \y(N)$ be an i.i.d. sample of a $d$-variate distribution, and let $\teta$ be non random vectors in a compact subset $\Theta\in\Rb^m$. Moreover, let
$h(\teta,\y)$ be a Borel measurable function on $\Rb^d\times \Theta$ such that for each $\y$, $h(\teta,\y)$ is a continuous function on $\Theta$. Finally, assume that
$\Eb[\sup_{\teta\in\Theta}|h(\teta,\x) |]<\infty$. Then we have almost surely
\be
\lim_{N\to\infty} \sup_{\teta\in\Theta}\Big\| \EE\big[h\big(\teta,\y\big)\big] - \Eb\big[h(\teta,\y)\big]  \Big\|=0.
\ee
\end{theo}
\begin{lemm}\label{USLLN2}
Let $G:\Rb\to\Rb$ be a nonlinearity function and $\x$ be the standardized signal. Suppose that Assumptions
(1)-(3) in Section \ref{assumption} are satisfied. Then we have
\be
\sup_{\w\in\Sc} \|\Eb[G(\w^\TR\x)] - \EE[G(\w^\TR\x)]\|\cas 0.
\ee
\end{lemm}
\begin{IEEEproof}
It suffices to show
\ben\label{equation75a}
\sup_{\w\in\Sc} \Big\|\Eb\Big[G_i(\w^\TR\A\s)\Big] - \frac{1}{N}\sum_{t=1}^NG_i\Big(\w^\TR\widehat{\A} \s(t)\Big)\Big\| \nonumber \\
\cas 0.
\een
It is easily seen that $\hata\cas\A$.
 Besides
by hypothesis of $G$ there holds
\be
\Eb\Big[\sup_{\w\in\Sc}|G(\w^\TR\x)|\Big]
& \leq & \Eb\Big[\sup_{\w\in\Sc}c(|\w^\TR\x|^p+1)\Big] <\infty.
\ee
Note that $\Sc$ is a compact set.
Applying USLLN to $G(\w^\TR\A\s)$ gives
\ben
\sup_{\w\in\Sc} \Big\|\Eb[G(\w^\TR\A\s)] - \frac{1}{N}\sum_{t=1}^NG\Big(\w^\TR\A \s(t)\Big)\Big\| \nonumber \\
\cas 0. \label{equation75b}
\een
Next, let us show that
\ben
\sup_{\w\in\Sc} \Big\|\frac{1}{N}\sum_{t=1}^NG\Big(\w^\TR\widehat{\A} \s(t)\Big) - \frac{1}{N}\sum_{t=1}^NG\Big(\w^\TR\A \s(t)\Big)\Big\|
\nonumber \\
\cas 0. \quad\quad \label{equation75c}
\een
Using the mean value theorem, it is easily seen that the term on the left hand side above is bounded by
\be
\frac{c}{N}\sum_{t=1}^N \Big(\|\hatc \s(t)\|+ \|\A \s(t) \|\Big)^p \|\s(t)\| \|\hata -\A \|,
\ee
which converges to zero almost surely.
\end{IEEEproof}
Lemma \ref{definite} and Lemma \ref{proposition627a} below can be found in e.g. \cite{ERDO}. Here we state them without proofs.
\begin{lemm}\label{definite}
For a  full rank square matrix $\M$, we have $(\M\M^\TR)^{-1/2}\M=\I$
if and only if $\M$ is symmetric and positive definite.
\end{lemm}
\begin{lemm}\label{proposition627a}
An orthogonal matrix $\W$ is a fixed point of $\Fc$ (resp. $\widehat{\Fc}$) if and only if there exists a diagonal matrix $\Lbd$ verifying $\Lbd^2=\I$, such that
$\Hc(\W)\W^{\TR}\Lbd$ (resp. $\widehat{\Hc}(\W)\W^{\TR}\Lbd$)
is symmetric and positive definite.
\end{lemm}
\begin{lemm} \label{lemma627a}
Let $\W^*$ be a given demixing matrix. There exists $r>0$ such that for any $\W\in\Bs_r(\W^*)\cap\Ob^d$, matrix
$\widehat{\Hc}(\W)\W^{\TR}\Lbd^*$
 is almost surely positive definite  provided that $N$ is large enough, where
 \be
 \Lbd^*\defeq\diag\big(\sign(\alpha_1),\ldots,\sign(\alpha_d)\big).
 \ee
\end{lemm}
\begin{IEEEproof}
First, we show that if $\R=\diag(r_1,\ldots,r_d)$ is a diagonal matrix with strictly positive diagonal entries $r_i$,
 then for any perturbation $\Dta$ such that $\|\Dta\|<r_k \defeq \min_{i} \{r_i\}$, the matrix $\R+\Dta$ is positive definite. In fact, for any
 $\x$, we have $\x^\TR(\R+\Dta)\x=\x^\TR\R\x + \x^\TR\Dta\x$, where
 \be
 |\x^\TR\Dta\x|\leq \|\Dta\|\|\x\|^2<r_k\|\x\|^2 < \x^\TR\R\x.
 \ee
This means $\x^\TR(\R+\Dta)\x>0$, hence $\R+\Dta$ is positive definite.

Now let us denote
\be
\Kc(\W)\defeq\diag\big(\gg'(\W\x)\big)\Lbd^* - g(\W\x)\x^\TR\W^{\TR}\Lbd^*.
\ee
Then we have
\be
\Eb[\Kc(\W)]
={\Hc}(\W)\W^{\TR}\Lbd^*.
\ee
As shown in the proof of Theorem \ref{theorem1} (i), there holds
\be
\Eb[\Kc(\W^*)]&=&{\Hc}(\W^*)\W^{*\TR}\Lbd =\L\W^*\W^{*\TR}\Lbd^* \\
&=&\diag(|\alpha_i|),
\ee
Denote $\epsilon=\min_{i}\{|\alpha_i|\}$.
By the continuity of $\Kc$, there exists $r$ such that
\ben\label{equation76a}
\sup_{\W\in\Bs_r(\W^*)}\|\Eb[\Kc(\W^*)] - \Eb[\Kc(\W)] \|<\frac{\epsilon}{2}.
\een
Besides, applying USLLN gives
\be
\sup_{\W\in\Bs_r(\W^*)}\| \Eb[\Kc(\W)] - \EE[\Kc(\W)] \|\cas 0,
\ee
If $N$ is large enough, then almost surely
\ben\label{equation76b}
\sup_{\W\in\Bs_r(\W^*)}\| \Eb[\Kc(\W)] - \EE[\Kc(\W)] \|\leq \frac{\epsilon}{2}.
\een
Combining (\ref{equation76a}) and (\ref{equation76b}) yields
\ben\label{perturbation}
\sup_{\W\in\Bs_r(\W^*)}\|\Eb[\Kc(\W^*)] - \EE[\Kc(\W)]  \|<\epsilon.
\een
Now that $\Eb[\Kc(\W^*)]$ is a diagonal matrix with strictly positive diagonal entries and the perturbation (\ref{perturbation})
  can be arbitrarily small, we conclude that
$\widehat{\Hc}(\W)\W^{\TR}\Lbd^*=\EE[\Kc(\W)]$ is almost surely positive definite for any $\W\in\Bs_r(\W^*)$.
\end{IEEEproof}

\subsection{Proof of Theorem (\ref{theorem1}) (ii) \label{201561a}}
Using Lemma \ref{USLLN2}, we can show that
\be
 \sup_{\W\in\Bs_r(\W^*)}\|\Jc_{\W^*}(\W)-\widehat{\Jc}_{\W^*}(\W)\|\cas 0
\ee
for any $r>0$. Hence there exists a local minimizer of
 \ben\label{218a}
\widehat{\Jc}_{\W^*}(\W)
&=& \sum_{i=1}^d \sign(\alpha_{i})\EE[G_i(\w_i^\TR{\x})]
\een
 in $\Bs_r(\W^*)$ on $\Ob(d)$.
  Let us denote this local minimizer by $\widehat{\W}$.  Now we show that $\widehat{\W}$ is also a fixed point of $\widehat{\Fc}$.
 By Lemma \ref{proposition627a}, it suffices to find a diagonal matrix $\Lbd$ such that $\Lbd^2=\I$ and
 \be
 \widehat{\Hc}(\widehat{\W})\widehat{\W}^{\TR}\Lbd=\EE\Big[\diag\Big(\gg'(\hatw\x)\Big) - \gg(\hatw\x)\x^\TR\hatw^{\TR}\Big]\Lbd
 \ee
 is a symmetric and positive definite matrix.

 We take $\Lbd=\Lbd^*=\diag\big(\sign(\alpha_1),\ldots,\sign(\alpha_d)\big)$.
  The positiveness of  $\widehat{\Hc}(\widehat{\W})\widehat{\W}^{\TR}\Lbd^*$ is confirmed by Lemma \ref{lemma627a}.
  Then it remains to prove the symmetry. Since $\diag\big(\gg'(\hatw\x)\big)$ is symmetric, we need only to show that $\EE\Big[\gg(\hatw\x)\x^\TR\hatw^{\TR}\Big]\Lbd^*$ is also symmetric.

Now let us return to the minimization of (\ref{218a}).  Notice that $G_1,\ldots,G_d$ are all even functions, therefore
not only $\widehat{\W}$, but also $\U\defeq\Lbd^*\widehat{\W}$ are  local minimizers of $\widehat{\Jc}_{\W^*}$ on $\Ob(d)$.
 Now write
\be
\widehat{\Jc}_{\W^*}(\W)&=& \sum_{i=1}^d \EE[\tilde{G}_i(\w_i^\TR{\x})],
\ee
 where $\tilde{G}_i=\sign(\alpha_i)G_i$ for $i=1,\ldots,d$.
Applying Lemma \ref{lemma626a} to $\U$ and $\widehat{\Jc}_{\W^*}$, we obtain
 the symmetry of
 \be
 \EE[\tilde{\gg}({\U}\x)\x^\TR{\U}^\TR]=\EE[\Lbd{\gg}(\Lbd\widehat{\W}\x)\x^\TR\widehat{\W}^\TR\Lbd].
\ee
Since $G_1,\ldots,G_d$ are all even functions, $g_1,\ldots,g_d$ are odd. It follows that matrix
\be
\EE[\Lbd{\gg}(\Lbd\widehat{\W}\x)\x^\TR\widehat{\W}^\TR\Lbd]=\EE[{\gg}(\widehat{\W}\x)\x^\TR\widehat{\W}^\TR\Lbd]
\ee
is symmetric. The proof is then achieved.
\section{Proof of Lemma \ref{estimating_equation} \label{proof_estimating}}
 To achieve Lemma \ref{estimating_equation}, we rely on Lemma \ref{optimum_condition}, which states that if an orthogonal matrix $\widehat{\W}$ optimizes
\be
\sum_{i=1}^d \EE\sign(\alpha_i)[G_i(\w_i^\TR\x)] \defeq \sum_{i=1}^d \EE[\tilde{G}_i(\w_i^\TR\x)],
\ee
 then it is such that $\EE[\tilde{\gg}(\widehat{\W}\x)\x^\TR\widehat{\W}^\TR]$ is symmetric.
Taking into account the orthogonality constraint, we derive the following characterization: $\widehat{\W}$ must satisfy
\ben
\hatw^\TR\hatw&=&\I,  \label{wtw}\\
\EE[\tilde{\gg}(\hatw\x)\x^\TR\hatw^\TR]&=&\EE[\hatw\x\tilde{\gg}(\hatw\x)^\TR]. \nonumber
\een
Applying the change of variable $\widehat{\B}\defeq \widehat{\W}\hatc^{-1/2}$ yields
\ben
\hatb^\TR\hatc\hatb&=&\I, \label{bcb}\\
\EE[\tilde{\gg}(\hatb\hatc^{1/2}\x)\x^\TR\hatc^{1/2}\hatb^\TR]&=&\EE[\hatb\hatc^{1/2}\x\tilde{\gg}(\hatb\hatc^{1/2}\x)^\TR]. \nonumber\\
\label{218b}
\een
Introducing auxiliary variable $\hat{\muu}=\bar{\y}$ and recalling that $\hatc=\Eb_{\y}[(\y-\bar{\y})(\y-\bar{\y})^\TR]$, we can rewrite (\ref{bcb}) as
\ben \label{218c}
\Eb_{\y}[\hatb^\TR(\y-\hat{\muu})(\y-\hat{\muu})^\TR\hatb]&=&\I
\een
Besides, substituting $\x(t)=\hatc^{-1/2}(\y(t)-\hat{\muu})$ in (\ref{218b}) gives
\ben
\EE\big[\tilde{\gg}\big(\hatb(\y-\hat{\muu})\big)(\y-\hat{\muu})^\TR\hatb^\TR\big] \nonumber \\
=\EE\big[\hatb(\y-\hat{\muu})\tilde{\gg}\big(\hatb(\y-\hat{\muu})\big)^\TR\big].
\label{218d}
\een
Combining (\ref{218c}) (\ref{218d}) and the auxiliary constraint $\muu=\bar{\y}$ together, we get
$\EE[\PSI(\hat{\teta},\y)]=0$.  Statement (ii) of Lemma \ref{estimating_equation} follows from a similar argument.

\section{Proof of Theorem \ref{main2} \label{proof_main2}}
We give the proof for $\B=\H^{-1}$. In this case, $\b_i$ corresponds to the extraction of $s_i$ for $i=1,\ldots, d$ and the permutation $\sigma$ appeared in Theorem \ref{main2} is an identity permutation.
The general case can be treated similarly.

According to \cite{VAAR}, the asymptotic normality of the estimator relies on the following conditions:
\begin{itemize}
\item[-] for every $\teta_1$
and $\teta_2$ in a neighborhood of $\teta^*$, there exists a measurable function $K(\cdot)$ with $\Eb[K(\y)^2]<\infty$ such that
\be
\|\PSI(\teta_1,\y)-\PSI(\teta_2,\y)\|\leq K(\y)\|\teta_1-\teta_2\|;
\ee
\item[-] $\Eb[\|\PSI(\teta^*,\y)\|^2]<\infty$;
\item[-] the map $\teta\rightarrow \Eb[\PSI(\teta,\y)]$ is differentiable at a zero $\teta^*$;
\item[-] $\hat{\teta}\xrightarrow{\Pb}\teta^*$;
\end{itemize}
With the assumptions stated in Section \ref{assumption}, it is not very hard to verify these conditions.

 Now, we are going to solve (\ref{equation2}) for $\T$.
 Write
\be
\T&=&(\t_{\muu}^\TR, \t_{11} , \t_{12}, \ldots ,\t_{dd}, \t_{\w_{1}\w_{2}}^\TR, \ldots,\t_{\w_{d-1}\w_{d}}^\TR)^\TR ,\\
\M&=&(\m_{\muu}^\TR, \m_{11} , \m_{12}, \ldots ,\m_{dd}, \m_{\w_{1}\w_{2}}^\TR, \ldots,\m_{\w_{d-1}\w_{d}}^\TR)^\TR ,\\
\Q&=&\Eb
\begin{bmatrix}
\partial_{\w_1}\PSI_{\muu}  & \cdots & \partial_{\w_d}\PSI_{\muu} &\partial_{\muu}\PSI_{\muu} \\
\partial_{\w_1}\PSI_{11}  & \cdots & \partial_{\w_d}\PSI_{11} &\partial_{\muu}\PSI_{11} \\
\vdots & \ddots & \vdots & \vdots \\
\partial_{\w_1}\PSI_{\w_{d-1}\w_d}  & \cdots & \partial_{\w_d}\PSI_{\w_{d-1}\w_d} &\partial_{\muu}\PSI_{\w_{d-1}\w_d} \\
\end{bmatrix},
\ee
where
\be
\partial_{\w_1}\PSI_{\muu} \defeq \frac{\partial}{\partial{\w_1}}\PSI_{\muu}(\teta,\y)\Big|_{\teta=\teta^*} .
\ee
The entries of $\Q$ can be given explicitly:
\be
\Eb[\partial_{\w_k} \PSI_{ij}]  & = & 0, \quad k\neq i,j\\
\Eb[\partial_{\w_i} \PSI_{ij}] & = & \h_j^\TR \\
\Eb[\partial_{\w_i} \PSI_{ii}] & = & 2\h_i^\TR \\
\Eb[\partial_{\muu} \PSI_{ij}] & = & 0 \\
\Eb[\partial_{\w_k} \PSI_{\w_i\w_j}] & = & 0, \quad k\neq i,j \\
\Eb[\partial_{\w_i} \PSI_{\w_i\w_j}] & = & \lambda_{ij}\h_j^\TR \\
\Eb[\partial_{\muu} \PSI_{\w_i\w_j}] & = & \eta_j\b^\TR_i - \eta_i\b^\TR_j, \quad j\neq i,
\ee
where $\lambda_{ij}=\Eb[g'(z_i)   - g(z_j) z_j]$.

It then follows that
\be
\t_{\w_i}&=& \sum_{j\neq i}^d\frac{\b_j\m_{\w_i\w_j} }{\lambda_{ij}+\lambda_{ji}}
+\sum_{j\neq i}^d\frac{\b_j\lambda_{ji}\m_{ij}}{\lambda_{ij}+\lambda_{ji}}\\
&&
+ \sum_{j\neq i}^d \frac{ (\eta_j\b_j\b_i^\TR - \eta_i\b_j\b_j^\TR)\m_{\muu}}{\lambda_{ij}+\lambda_{ji}}
+ \frac{\b_i\m_{ii}}{2},
\ee

Next, we calculate $\Eb[\t_{\w_i}\t_{\w_i}^\TR]$, which is the asymptotic covariance matrix of $N^{1/2}(\hat{\b}_i - \b_i)$. To achieve this,
we need to compute $\Eb[\PSI(\teta^*,\y) \PSI(\teta^*,\y)^\TR]$ first, which is the covariance matrix of $\M$.
Let $i,j,k,l$ be different subscripts.
It is easily seen that
\be
\Eb[\PSI_{\muu}\PSI^{\TR}_{\muu}]&=&\H\H^\TR\\
\Eb[\PSI_{\muu}\PSI_{ij}]&=& 0 \\
\Eb[\PSI_{\muu}\PSI_{ii}]&=&\h_i \Eb[s_i^3]  \\
\Eb[\PSI_{\muu}\PSI_{\w_i\w_j}]&=&  \eta_i\h_j - \eta_j\h_i \\
\Eb[\PSI_{ii} \PSI_{ii} ] & = & \tau_i \\
\Eb[\PSI_{ij} \PSI_{ji} ] & = & 1 \\
\Eb[\PSI_{ij} \PSI_{il} ] & = & 0 \\
\Eb[\PSI_{ii} \PSI_{ij} ] & = & 0 \\
\Eb[\PSI_{\w_i\w_j} \PSI_{\w_i\w_j} ]
& = & \beta_i + \beta_j - 2\gamma_i\gamma_j ,\\
\Eb[\PSI_{\w_i\w_j} \PSI_{\w_i\w_k} ] & = &  \eta_j\eta_k, \\
\Eb[\PSI_{\w_i\w_j} \PSI_{\w_j\w_k} ] & = & -\eta_j\eta_k, \\
\Eb[\PSI_{\w_i\w_j} \PSI_{\w_k\w_j} ] & = &  \eta_j\eta_k, \\
\Eb[\PSI_{\w_i\w_j} \PSI_{\w_k\w_i} ] & = & -\eta_j\eta_k,  \\
\Eb[\PSI_{\w_i\w_j} \PSI_{\w_k\w_l} ] & = & 0 \\
\Eb[\PSI_{ii} \PSI_{\w_i\w_j}] & = & -\Eb[s_i^3]\eta_j\\
\Eb[\PSI_{ii} \PSI_{\w_j\w_k}] & = & 0 \\
\Eb[\PSI_{ij} \PSI_{\w_i\w_j}] & = & \gamma_i - \gamma_j \\
\Eb[\PSI_{ik} \PSI_{\w_i\w_j}] & = & 0 \\
\Eb[\PSI_{kl} \PSI_{\w_i\w_j}] & = & 0.
\ee
After some tedious algebraic simplifications,  expression (\ref{ASYM2}) follows.

\section{Some probability distributions \label{somePDF}}
\subsection{Generalized Gaussian distribution $GG(\alpha)$  \label{GG}}
The generalized Gaussian density function with parameter $\alpha$, zero mean and unit variance is given by
\be
f_{\alpha}(x)=\frac{\alpha\beta_{\alpha}}{2\Gamma(1/\alpha)}\exp{\{-(\beta_{\alpha}|x|)^{\alpha}\}},
\ee
where $\alpha$ is a positive parameter that controls the distribution¡¯s exponential rate of decay, $\Gamma$ is the Gamma function, and
\be
\beta_{\alpha}=\sqrt{\frac{\Gamma(3/\alpha)}{\Gamma(1/\alpha)}}.
\ee
This generalized Gaussian family encompasses the ordinary
standard normal distribution for $\alpha=2$ , the Laplace distribution for $\alpha=1$, and the uniform distribution in the limit $\alpha\to\infty$.
\subsection{Bimodal distribution with Gaussian mixture\label{sectionbimod}}
The bimodal distribution used in this paper consists of a mixture of two Gaussian distribution. Define random variable
\be
X=Z Y_1 + (1-Z)Y_2,
\ee where
$Y_i\sim\Nc(\mu_i,\sigma_i^2)$ and $Z\sim\Bc(p)$ are mutually independent random variables.
 Here, $\Bc(p)$ denotes the Bernoulli distribution with parameter $p$, i.e. $\Pb(Z=1)=p$ and $\Pb(Z=0)=1-p$.
It is easy to see that the probability density function (PDF) of $X$ is given by
\be
f_X(x)= p f_{Y_1}(x) + (1-p)f_{Y_2}(x),
\ee
where $f_{Y_i}$ is the PDF of $Y_i$ for $i=1,2$.

Now take any $\mu_1,\mu_2$ such that $\mu_1\mu_2<0$ and $|\mu_1\mu_2|<1$, then let $\sigma^2_1=\sigma^2_2=1-|\mu_1\mu_2|$
and
\be
p =\frac{|\mu_2|}{|\mu_1|+|\mu_2|}.
\ee
Defined in such a way, $X$ is a random variable with zero mean, unit variance and two modes at $\mu_1$ and $\mu_2$.
Notably, if $\mu_1\neq -\mu_2$, then the distribution of $X$ is asymmetric.
 Since the PDF of $X$ is completely determined by $\mu_1,\mu_2$, we use them as controlling parameter and denote by
 ``Bimod$(\mu_1,\mu_2)$'' the distribution of $X$.


\section*{Acknowledgement}
The author would also like to thank the editor and the anonymous referees
for carefully reading the manuscript and for giving us many
helpful and constructive suggestions resulting in the present
work.

\bibliographystyle{IEEEtran}
\bibliography{IEEEabrv,IeeeBibJul03}

\begin{thebibliography}{10}
\providecommand{\url}[1]{#1}
\csname url@samestyle\endcsname
\providecommand{\newblock}{\relax}
\providecommand{\bibinfo}[2]{#2}
\providecommand{\BIBentrySTDinterwordspacing}{\spaceskip=0pt\relax}
\providecommand{\BIBentryALTinterwordstretchfactor}{4}
\providecommand{\BIBentryALTinterwordspacing}{\spaceskip=\fontdimen2\font plus
\BIBentryALTinterwordstretchfactor\fontdimen3\font minus
  \fontdimen4\font\relax}
\providecommand{\BIBforeignlanguage}[2]{{%
\expandafter\ifx\csname l@#1\endcsname\relax
\typeout{** WARNING: IEEEtran.bst: No hyphenation pattern has been}%
\typeout{** loaded for the language `#1'. Using the pattern for}%
\typeout{** the default language instead.}%
\else
\language=\csname l@#1\endcsname
\fi
#2}}
\providecommand{\BIBdecl}{\relax}
\BIBdecl

\bibitem{WEISSP2}
T.~Wei, ``Asymptotic analysis of the generalized symmetric {FastICA}
  algorithm,'' in \emph{2014 IEEE Workshop on Statistical Signal Processing
  (SSP) (SSP'14)}, Gold Coast, Australia, Jun. 2014, pp. 484--487.

\bibitem{HYVABOOK}
A.~Hyv{\"a}rinen, J.~Karhunen, and E.~Oja, \emph{Independent Component
  Analysis}.\hskip 1em plus 0.5em minus 0.4em\relax New York:
  Wiley-Interscience, 2001.

\bibitem{COMOBOOK}
P.~Comon and C.~Jutten, \emph{Handbook of Blind Source Separation: Independent
  Component Analysis and Applications}.\hskip 1em plus 0.5em minus 0.4em\relax
  Academic Press, 2010, pp. 179--227.

\bibitem{AMABOOK}
S.-I. Amari and A.~Cichocki, \emph{Adaptive Blind Signal and Image
  Processing}.\hskip 1em plus 0.5em minus 0.4em\relax New York: Wiley, 2002.

\bibitem{CARD1993}
J.~F. Cardoso and A.~Souloumiac, ``Blind beamforming for non-gaussian
  signals,'' \emph{IEEE Proceedings-F}, vol. 140, no.~6, pp. 362--370, Dec.
  1993.

\bibitem{COMO94}
P.~Comon, ``Independent component analysis: a new concept?'' \emph{Signal
  Processing}, vol.~36, no.~3, pp. 287--314, Apr. 1994.

\bibitem{RADICAL}
E.~G. Learned-Miller and J.~W. Fisher, ``{ICA} using spacings estimates of
  entropy,'' \emph{Journal of machine learning research}, no.~4, pp.
  1271--1295, 2003.

\bibitem{ZARZ2010}
V.~Zarzoso and P.~Comon, ``Robust independent component analysis by iterative
  maximization of the kurtosis contrast with algebraic optimal step size,''
  \emph{Neural Networks, IEEE {T}ransactions on}, vol.~21, no.~2, pp. 248--261,
  Feb 2010.

\bibitem{HYVAOJA97}
A.~Hyv{\"a}rinen and E.~Oja, ``A fast fixed-point algorithm for independent
  component analysis,'' \emph{Neural Computation}, vol.~9, no.~7, pp.
  1483--1492, 1997.

\bibitem{HYVA99}
A.~Hyv{\"a}rinen, ``Fast and robust fixed-point algorithms for independent
  component analysis,'' \emph{Neural Networks, IEEE {T}ransactions on},
  vol.~10, no.~3, pp. 626--634, May 1999.

\bibitem{DELF}
N.~Delfosse and P.~Loubaton, ``Adaptive blind separation of independent
  sources: A deflation approach,'' \emph{Signal Processing}, vol.~45, no.~1,
  pp. 59 -- 83, 1995.

\bibitem{OLLI}
E.~Ollila, ``The deflation-based {FastICA} estimator: Statistical analysis
  revisited,'' \emph{Signal Processing, IEEE {T}ransactions on}, vol.~58,
  no.~3, Mar. 2010.

\bibitem{MIET2014a}
J.~Miettinen, K.~Nordhausen, H.~Oja, and S.~Taskinen, ``Deflation-based
  {FastICA} with adaptive choices of nonlinearities,'' \emph{Signal Processing,
  IEEE {T}ransactions on}, vol.~62, no.~21, pp. 5716--5724, Nov 2014.

\bibitem{OJAYUAN}
E.~Oja and Z.~Yuan, ``The {FastICA} algorithm revisited: Convergence
  analysis,'' \emph{Neural Networks, IEEE {T}ransactions on}, vol.~17, no.~6,
  2006.

\bibitem{TICHOJA2}
Z.~Koldovsky, P.~Tichavsky, and E.~Oja, ``Efficient variant of algorithm
  {FastICA} for independent component analysis attaining the {Cram\'er-Rao}
  lower bound,'' \emph{Neural Networks, IEEE {T}ransactions on}, vol.~17,
  no.~5, pp. 1265--1277, 2006.

\bibitem{TICHOJA}
P.~Tichavsky, Z.~Koldovsky, and E.~Oja, ``Performance analysis of the {FastICA}
  algorithm and {Cram\'er-Rao} bounds for linear independent component
  analysis,'' \emph{Signal Processing, IEEE {T}ransactions on}, vol.~54, no.~4,
  pp. 1189--1203, Apr. 2006.

\bibitem{OLLI2008}
E.~Ollila, H.-J. Kim, and V.~Koivunen, ``Compact {Cram\'er-Rao} bound
  expression for independent component analysis,'' \emph{Signal Processing,
  IEEE {T}ransactions on}, vol.~56, no.~4, pp. 1421--1428, 2008.

\bibitem{SHEN}
H.~Shen, M.~Kleinsteuber, and K.~H{\"{u}}per, ``Local convergence analysis of
  {FastICA} and related algorithms,'' \emph{Neural Network, IEEE {T}ransactions
  on}, vol.~19, no.~6, pp. 1022--1032, Jun. 2008.

\bibitem{REGA}
P.~A. Regalia and E.~Kofidis, ``Monotonic convergence of fixed-point algorithms
  for {ICA},'' \emph{Neural Network, IEEE {T}ransactions on}, vol.~14, no.~4,
  pp. 943--949, Jul. 2003.

\bibitem{DOUG2003}
S.~Douglas, ``On the convergence behavior of the {FastICA} algorithm,'' in
  \emph{Proc. 4th Symp. Independent Component Analysis Blind Source
  Separation}, Nara, Japan, apr 2003, pp. 409--414.

\bibitem{HYVABING}
A.~Hyv{\"a}rinen and E.~Bingham, ``A fast fixed-point algorithm for independent
  component analysis of complex-valued signals,'' \emph{Int. J. Neural Syst.},
  vol.~10, no.~1, pp. 1--8, 2000.

\bibitem{HYVA97}
A.~Hyv{\"a}rinen, ``One-unit contrast functions for independent component
  analysis: A statistical analysis,'' in \emph{Proc. IEEE NNSP Workshop
  '97}.\hskip 1em plus 0.5em minus 0.4em\relax Neural Networks for Signal
  Processing VII, 1997.

\bibitem{OLLI2011}
K.~Nordhausen, P.~Ilmonen, A.~Mandal, H.~Oja, and E.~Ollila, ``Deflation-based
  {FastICA} reloaded,'' in \emph{19th European Signal Processing Conference
  (EUSIPCO 2011)}, Barcelona, Spain, Sep. 2011.

\bibitem{REYH2012}
N.~Reyhani, J.~Ylipaavalniemi, R.~Vigario, and E.~Oja, ``Consistency and
  asymptotic normality of {FastICA} and bootstrap {FastICA},'' \emph{Signal
  Processing}, vol.~92, pp. 1767--1778, 2012.

\bibitem{WEI2}
A.~Dermoune and T.~Wei, ``{FastICA} algorithm: Five criteria for the optimal
  choice of the nonlinearity function,'' \emph{Signal Processing, IEEE
  {T}ransaction on}, vol.~61, no.~8, pp. 2078--2087, Apr. 2013.

\bibitem{HYVA2006}
A.~Shimizu, A.~Hyv{\"{a}}rinen, K.~Yutaka, P.~Hoyer, and A.~J. Kerminen,
  ``Testing signifcance of mixing and demixing coefficients in {ICA},'' in
  \emph{Int. Conf. Independent Component Analysis ({ICA} 2006)}, 2006.

\bibitem{HYVAOJA2000}
A.~Hyv{\"a}rinen and E.~Oja, ``Independent component analysis: Algorithms and
  applications,'' \emph{Neural Networks}, vol.~13, no. 4-5, pp. 411--430, 2000.

\bibitem{HYVAOJA98}
A.~Hyvarinen and E.~Oja., ``Independent component analysis by general
  non-linear {Heb}bian-like learning rules.'' \emph{Signal Processing},
  vol.~64, no.~3, pp. 301--313, 1998.

\bibitem{WEI3}
T.~Wei, ``On the fixed points and spurious solutions of the {FastICA} algorithm
  (in revision),'' \emph{Neural Computing and Applications}, 2015.

\bibitem{VAAR}
A.~{van der Vaart}, \emph{Asymptotic Statistics}.\hskip 1em plus 0.5em minus
  0.4em\relax Cambridge University Press, 2000, ch.~5.

\bibitem{FIOR2006}
S.~Fiori, ``Fixed-point neural independent component analysis algorithms on the
  orthogonal group,'' \emph{Future generation computer system}, vol.~22, pp.
  430--440, 2006.

\bibitem{ERDO}
A.~Erdogan, ``On the convergence of {ICA} algorithms with symmetric
  orthogonalization,'' \emph{Signal Processing, IEEE {T}ransactions on},
  vol.~57, no.~6, pp. 2209--2221, 2009.

\bibitem{ILMO2010a}
P.~Ilmonen, K.~Nordhausen, H.~Oja, and E.~Ollila, ``A new performance index for
  {ICA}: Properties, computation and asymptotic analysis,'' in \emph{Latent
  Variable Analysis and Signal Separation}, ser. Lecture Notes in Computer
  Science, V.~Vigneron, V.~Zarzoso, E.~Moreau, R.~Gribonval, and E.~Vincent,
  Eds.\hskip 1em plus 0.5em minus 0.4em\relax Springer Berlin Heidelberg, 2010,
  vol. 6365, pp. 229--236.

\bibitem{MIET2014b}
J.~{Miettinen}, S.~{Taskinen}, K.~{Nordhausen}, and H.~{Oja}, ``{Fourth Moments
  and Independent Component Analysis},'' \emph{ArXiv e-prints}, Jun. 2014.

\bibitem{VIRT2015}
J.~{Virta}, K.~{Nordhausen}, and H.~{Oja}, ``{Joint Use of Third and Fourth
  Cumulants in Independent Component Analysis},'' \emph{ArXiv e-prints}, May
  2015.

\bibitem{BIER}
H.~Bierens, \emph{Introduction to the Mathematical and Statistical Foundations
  of Econometrics}.\hskip 1em plus 0.5em minus 0.4em\relax Cambridge University
  Press, 2005, p. 171.

\bibitem{NEWE1991}
W.~K. Newey, ``Uniform convergence in probability and stochastic
  equicontinuity,'' \emph{Econometrica}, vol.~59, no.~4, pp. 1161--1167, Jul.
  1991.

\bibitem{ANDR1992}
D.~W.~K. Andrews, ``Generic uniform convergence,'' \emph{Econometric Theory},
  vol.~02, no.~9, Jun. 1992.

\end{thebibliography}

\begin{IEEEbiography}[{\includegraphics[width=1in,height=1.25in,clip,keepaspectratio]{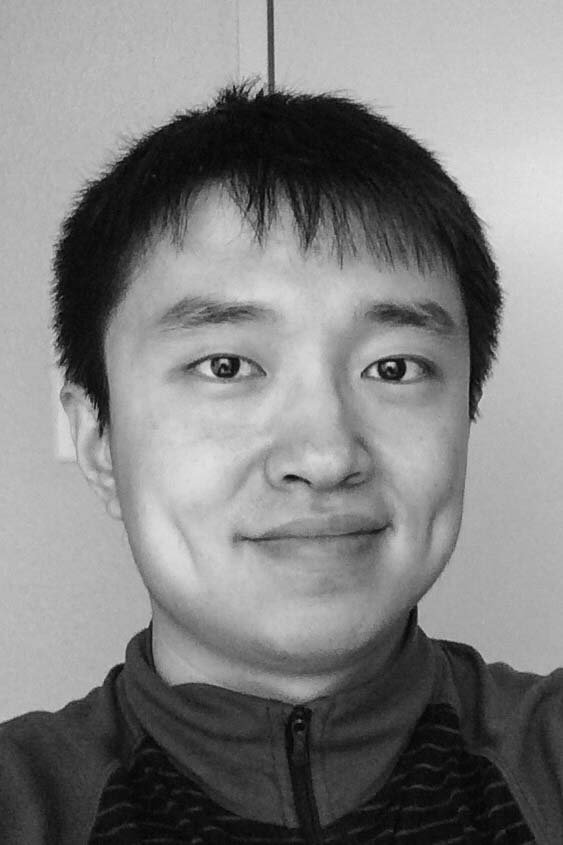}}]{Tianwen Wei} graduated from Wuhan University, Wuhan, China, in 2006. He received the M.S. degree in 2009 and Ph.D. in 2013 from the University of Sciences and Technology of Lille, Villeneuve d'Ascq, France.

From 2014 to 2015 he was a post-doctoral researcher
at the University of Franche-Comt\'e, Besan\c{c}on, France.
Currently he is employed as an assistant professor at the Department of Statistics and Mathematics,  Zhongnan University of Economics and Law, Wuhan, China.
His research interests include independent component analysis, convex optimization and machine learning.
\end{IEEEbiography}
\end{document}